\documentclass[nohyperref]{article}

\usepackage{microtype}
\usepackage{graphicx}
\usepackage{subfig}
\usepackage{enumitem}
\usepackage{booktabs} 
\usepackage[none]{hyphenat}

\usepackage{hyperref}

\usepackage[accepted]{icml2022}

\usepackage{amsmath}
\usepackage{amssymb}
\usepackage{mathtools}
\usepackage{amsthm}

\usepackage[capitalize,noabbrev]{cleveref}

\theoremstyle{plain}

\theoremstyle{definition}

\theoremstyle{remark}

\usepackage[disable,textsize=tiny]{todonotes}

\icmltitlerunning{Learning from Mistakes based on Class Weighting}

\begin{document}

\twocolumn[
\icmltitle{Learning from Mistakes based on Class Weighting with Application to Neural Architecture Search}



\icmlsetsymbol{equal}{*}

\begin{icmlauthorlist}
\icmlauthor{Jay Gala}{yyy}
\icmlauthor{Pengtao Xie}{comp}
\end{icmlauthorlist}

\icmlaffiliation{yyy}{Mumbai, India}
\icmlaffiliation{comp}{UC San Diego, USA}

\icmlcorrespondingauthor{Jay Gala}{jaygala24@gmail.com}
\icmlcorrespondingauthor{Pengtao Xie}{p1xie@eng.ucsd.edu}

\icmlkeywords{Machine Learning, Neural Architecture Search, DARTS, Class Weighting, Optimization}

\vskip 0.3in
]



\printAffiliationsAndNotice{}  

\begin{abstract}
Learning from mistakes is an effective learning approach widely used in human learning, where a learner pays greater focus on mistakes to circumvent them in the future to improve the overall learning outcomes. In this work, we aim to investigate how effectively we can leverage this exceptional learning ability to improve machine learning models. We propose a simple and effective multi-level optimization framework called learning from mistakes using class weighting (LFM-CW), inspired by mistake-driven learning to train better machine learning models. In this formulation, the primary objective is to train a model to perform effectively on target tasks by using a re-weighting technique. We learn the class weights by minimizing the validation loss of the model and re-train the model with the synthetic data from the image generator weighted by class-wise performance and real data. We apply our LFM-CW framework with differential architecture search methods on image classification datasets such as CIFAR and ImageNet, where the results show that our proposed strategy achieves lower error rate than the baselines.
\end{abstract}

\section{Introduction}
\label{sec:introduction}

\begin{figure}[ht]
    \vskip 0.2in
    \begin{center}
    \centerline{\includegraphics[width=0.9\columnwidth]{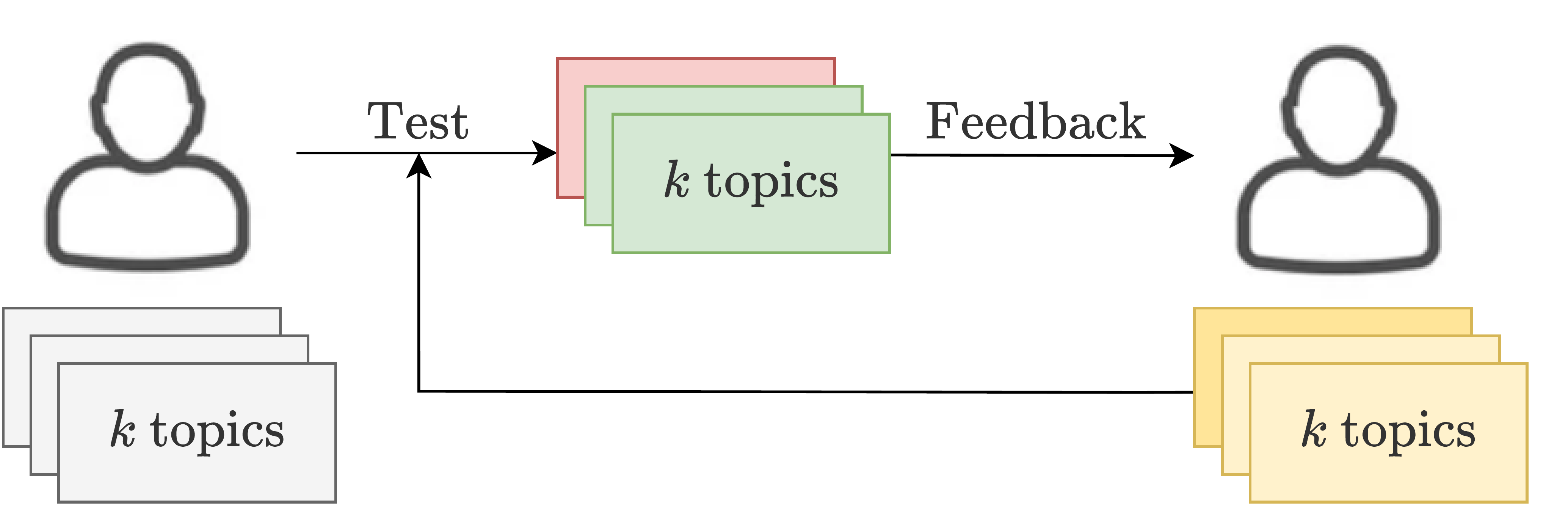}}
    \caption{Illustration of learning from mistakes. Learner takes a test after studying $k$ topics and receives feedback on topic-wise performance. Learner revisits the topics again, focusing on weaker topics (indicated by darker shade) in comparison to other topics (indicated by lighter shade).}
    \label{fig:lfm_main_fig}
    \end{center}
    \vskip -0.4in
\end{figure}

Learning from mistakes is an effective learning approach widely used in human learning. A student learns some topics from the textbook and then takes a test to evaluate how well he/she has grasped them. Topics on which the student made mistakes in the test were not thoroughly understood. The student devotes more attention on improving the understanding of these topics to prevent repeating similar mistakes in the future. Intuitively, this approach helps the student to gauge which concepts are weaker and encourages them to concentrate more on these in order to reinforce a better understanding.

We are interested in investigating whether we can utilize this human learning method to train better machine learning models. We propose a novel machine learning framework called learning from mistakes using class weighting (LFM-CW) (as illustrated in Figure \ref{fig:lfm_main_fig}), an extension to Skillearn \cite{Xie2020SkillearnML}. In this framework, the formulation consists of a three-level optimization problem that involves three learning stages. In this work, we assume the end task is classification while noting that our framework can also be extended to other tasks.

In this formulation, there is a classification model and a conditional image generator (CIG). The classification model has a learnable architecture with a set of network weights. CIG is a generative adversarial network (GAN) \cite{Goodfellow2014GenerativeAN}, where the generator has a fixed architecture with a set of learnable weights, and the discriminator has a learnable architecture and a set of learnable weights. We train a CIG, which takes a class label as an input and generates an image belonging to this class. After training an image classification model, we measure its class-wise validation performance. If the classification model does not perform in a certain class $c$, the CIG will generate more images belonging to $c$ and then use these generated images to re-train the classification model. The intuition is: if the
classification model does not perform well on a certain class $c$, we assign more weights to its training examples so as to enforce more attention of the classification model to this class.

A multi-level optimization framework is developed to unify the three learning stages and optimize them jointly in an end-to-end manner. Each learning stage influences other stages. We apply our method for neural architecture search in image classification tasks on CIFAR-10 \cite{Krizhevsky2009LearningML}, CIFAR-100 \cite{Krizhevsky2009LearningML}, and ImageNet \cite{Deng2009ImageNetAL}. Experimental results demonstrate competitive performance with existing differential architecture search approaches \cite{Liu2019DARTSDA,Liang2019DARTSID,Chen2019ProgressiveDA,Xu2020PCDARTSPC}.

To summarize, the contribution of our work is three-fold:
\vspace{-1em}
\begin{enumerate}[leftmargin=*]
    \setlength\itemsep{0em}  
     \item We propose a novel machine learning approach called learning from mistakes using class weighting (LFM-CW) by leveraging the mistake-driven learning technique of humans. In our approach, we present a formulation where the model uses its intermediate network weights to make predictions and then re-train network weights by adopting a re-weighting strategy to incorporate the corrective feedback, resulting in improved learning outcomes.
    \item We develop an efficient optimization algorithm to solve the LFM-CW strategy, a multi-level framework consisting of three learning stages.
    \item We apply the LFM-CW strategy for neural architecture search on various benchmarks, where the results demonstrate the effectiveness of our method.
\end{enumerate}

\section{Related Work}
\label{sec:related_work}

\subsection{Neural Architecture Search}

Neural Architecture Search (NAS) aims to automatically design and build high-performing neural architectures that can achieve the best performance on a specific task with minimal human intervention. Although existing NAS approaches use a hybrid search strategy to find an ideal candidate cell and stack multiple copies to build a larger network leveraging human expertise as seen in human-designed architectures (e.g., GoogleNet \cite{Szegedy2015GoingDW}, ResNet \cite{He2016DeepRL}, etc.), NAS aims to automate the architecture search procedure fully. Early NAS techniques \cite{Zoph2017NeuralAS,Baker2017DesigningNN,Pham2018EfficientNA,Zoph2018LearningTA} used reinforcement learning (RL), in which a policy network learns to produce high-quality architectures by minimizing validation loss as a reward. Following RL-based approaches, evolutionary algorithms \cite{Real2017LargeScaleEO,Liu2018HierarchicalRF,Real2019RegularizedEF} validated the feasibility of automatic architecture search achieving comparable results, where high-quality architectures produce offspring to replace low-quality architectures, and quality is measured using fitness scores. However, these approaches are computationally expensive and not feasible for researchers who lack sufficient computational resources. To address this issue, differentiable search methods \cite{Liu2019DARTSDA,Xie2019SNASSN,Cai2019ProxylessNASDN} have been proposed, which aims to accelerate the search for neural architecture by parametrizing architectures as differentiable functions and optimizing using gradient descent-based approaches. Several subsequent works such as P-DARTS \cite{Chen2019ProgressiveDA}, PC-DARTS \cite{Xu2020PCDARTSPC} have further improved differential NAS algorithms. P-DARTS \cite{Chen2019ProgressiveDA} progressively increases the depth of architecture during architecture search. PC-DARTS \cite{Xu2020PCDARTSPC} samples sub-architectures from super network to eliminate redundancy in the search process. Our work is closely related to a meta-learning approach GTN \cite{Such2020GenerativeTN} where a generative model is trained to produce synthetic examples and then uses these examples to search neural architectures. In contrast, our approach aims to search neural architecture by jointly optimizing on synthetic and real data in a single stage.

\subsection{Importance Weighting}

Weighting is a widely used technique in machine learning to improve robustness against training data bias. To deal with class-imbalanced datasets corresponding to distributional shifts, several sample weighting strategies (e.g., focal-loss \cite{Lin2017FocalLF}, class-balanced loss \cite{Cui2019ClassBalancedLB}) have been proposed, where larger weights are assigned to difficult or easily misclassified examples and lower weights are assigned to well-classified examples. Similarly, many approaches \cite{Axelrod2011DomainAV,Foster2010DiscriminativeIW,Jiang2007InstanceWF,Moore2010IntelligentSO,Ngiam2018DomainAT,Sivasankaran2017DiscriminativeIW} showed that selecting or re-weighting training examples improves overall performance. Bi-level optimization-based approaches \cite{Ren2018LearningTR,Shu2019MetaWeightNetLA,Wang2020OptimizingDU,Ren2020NotAU,Wang2020MetaSemiAM} learn data weights by minimizing validation performance of models trained using re-weighted data, in an end-to-end manner. \cite{Liu2021JustTT} proposed a simple two-stage approach to improve group robustness, with the first stage identifying minority cases and the second stage upweighting these examples. On the other hand, our approach tries to learn network weights in the second stage using real data and synthetic data weighted based on the class-wise validation performance of the network trained in the first stage on the real data.

\section{Problem Formulation}
\label{sec:formulation}

\begin{figure*}[ht]
    \vskip 0.2in
    \begin{center}
    \centerline{\includegraphics[width=0.9\linewidth]{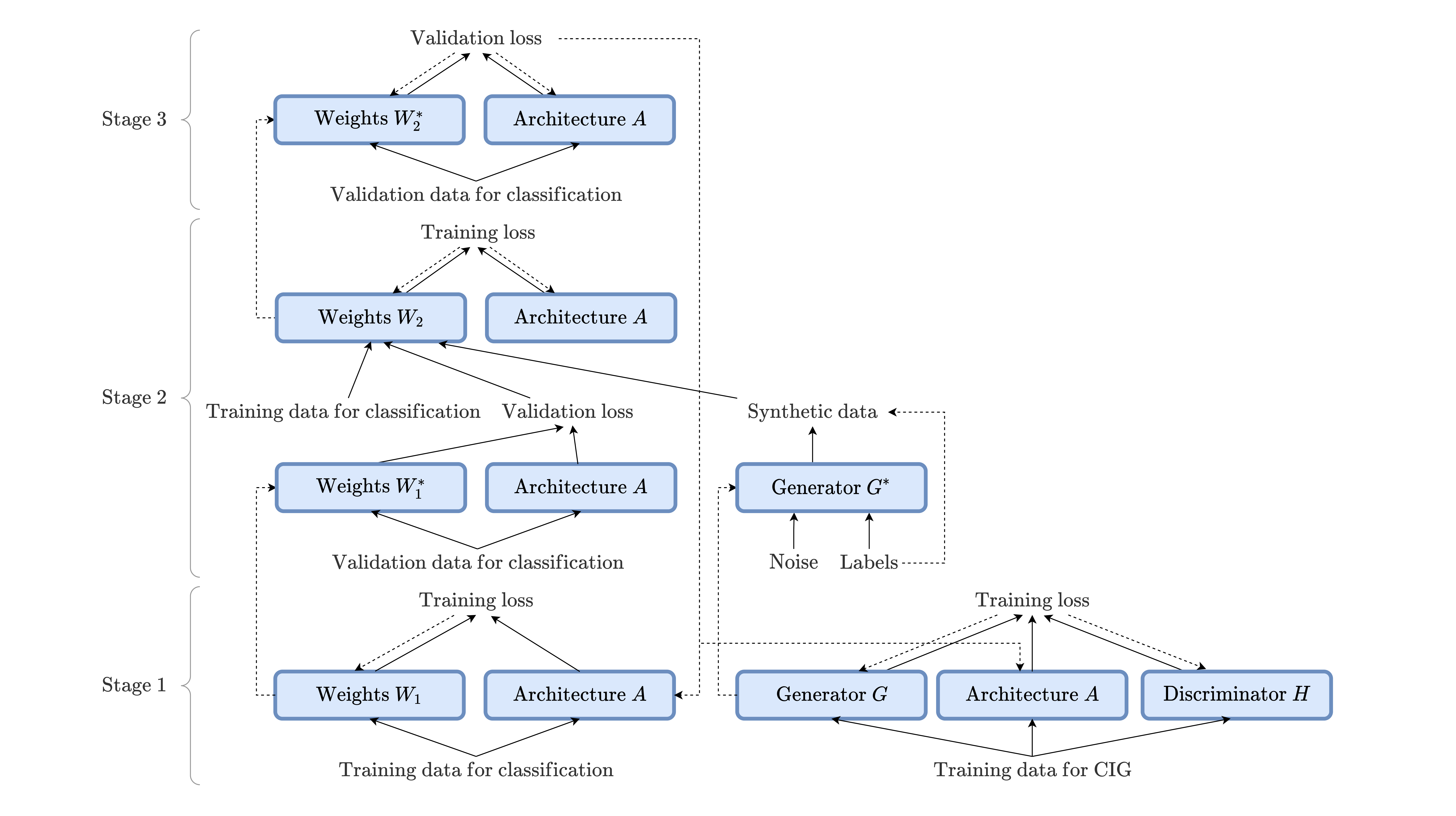}}
    \caption{Overview of LFM formulation. Solid arrows demonstrate the process of making predictions and calculating losses. Dotted arrows denote the process of updating learnable parameters by minimizing the corresponding losses.}
    \label{fig:f2_main_fig}
    \end{center}
    \vskip -0.2in
\end{figure*}

In this formulation (as displayed in Figure \ref{fig:f2_main_fig}), there is a classification model and GAN-based CIG (e.g., CGAN \cite{Mirza2014ConditionalGA}, SAGAN \cite{Zhang2019SelfAttentionGA}, BigGAN \cite{Brock2019LargeSG}, RoCGAN \cite{Chrysos2019RobustCG}) with a generator and a discriminator. The classification model has a learnable architecture $A$ and two sets of network weights $W_1$ and $W_2$. The generator has a fixed architecture and a set of learnable weights $G$. The discriminator has a learnable architecture, same as the architecture of the classification model, which is $A$ and a set of learnable weights $H$. Learning consists of three stages. We summarize the key elements of this formulation in Table \ref{tab:notation_f2}.

\begin{table}[h]
    \caption{Notations in LFM-CW framework}
    \label{tab:notation_f2}
    \vskip 0.15in
    \begin{center}
    \begin{small}
    \begin{tabular}{l|l}
    \hline
    Notation & Meaning \\
    \hline
    $A$ & Architecture \\
	$W_1$ & The first set of network weights \\
	$W_2$ & The second set of network weights \\
    $G$ & Network weights of Generator \\
    $H$ & Network weights of Discriminator \\
    $D_{cls}^{\mathrm(tr)}$ & Training dataset for classification \\
    $D_{cls}^{\mathrm(val)}$ & Validation dataset for classification \\
    $D_{cig}$ & Training dataset for CIG \\
    \hline
    \end{tabular}
    \end{small}
    \end{center}
    \vskip -0.1in
\end{table}

In the first stage, we train weights $W_1$ of the classification model on the training dataset $D_{cls}^{\mathrm(tr)} = \{(x_i, y_i)\}_{i=1}^{N}$, where $x_i$ is an image and $y_i$ is a class label, with architecture $A$ fixed as:

\begin{equation}
\label{eq:eq_1}
    W_1^*(A) = \min_{W_1} L(A, W_1, D_{cls}^{\mathrm(tr)})
\end{equation}

Meanwhile, we train a GAN-based CIG with a generator and a discriminator. The generator takes a class name as an input and generates an image. The discriminator takes an image as input and predicts whether it is synthetic or real. The training dataset for CIG is $D_{cig} = \{(y_i, x_i)\}_{i=1}^N$, which is obtained by switching the order of images and labels in $D_{cls}^{\mathrm(tr)}$. In this stage, when training the CIG, we train generator $G$ and discriminator $H$ with architecture $A$ fixed in the following way:

\begin{equation}
\label{eq:eq_2}
    G^*(A), H^*(A) = \min_{G} \max_{H} L(G, H, A, D_{cig})
\end{equation}

There are two optimization problems in this stage. In the second stage, we measure the validation performance of the optimally trained $W_1^*(A)$ on a validation set $D_{cls}^{\mathrm(val)}$. Let $l_c(A, W_1^*(A), D_{cls}^{\mathrm(val)})$ denotes the validation loss on class $c$. The smaller, the better. Meanwhile, we use the CIG to generate synthetic images. For each class $c$, we generate $M$ images $\{\hat{x_{c,m}}_{m=1}^{M}\}$, where $\hat{x_{c,m}} = f(c, \delta_m, G^*(A))$: the generator $f$ parameterized by $G^*(A)$ takes the class name $c$ and a random noise vector ${\delta}$ as inputs and generates $\hat{x_{c,m}}$. We use $\{\hat{x_{c,m}}_{m=1}^{M}\}$ to train the second set of weights $W_2$ of the classification model. These synthetic training examples are weighted using validation losses: if the validation loss on class $c$ is large, synthetic examples in class $c$ are given larger weights. In this stage, the optimization problem is:

\begin{multline}
\label{eq:eq_3}
    W_2^*(A, W_1^*(A), G^*(A)) = \min_{W_2} L(A, W_2, D_{cls}^{\mathrm(tr)})\ + \\ \hspace{2em} \lambda \sum_{c=1}^{C} l_c(A, W_1^*(A), D_{cls}^{\mathrm(val)}) \sum_{m=1}^{M} L(A, W_2, \hat{x_{c,m}}, c)
\end{multline}

where $\lambda$ is a trade-off parameter, $C$ is the total number of classes, $\hat{x_{c,m}}$ is the $m^{th}$ generated image when feeding class $c$ to the CIG model. In the third stage, we validate optimal weights $W_2^*(A, W_1^*(A), G^*(A))$ on the validation dataset $D_{cls}^{\mathrm(val)}$ and learn $A$ by minimizing the validation loss as:

\begin{equation}
\label{eq:eq_4}
    \min_{A} L(A, W_2^*(A, W_1^*(A), G^*(A)), D_{cls}^{\mathrm(val)})
\end{equation}

Putting the above pieces together, we have the following overall formulation.

\begin{equation}
\label{eq:eq_5}
    \begin{aligned}
        \min_{A}\ & L(A, W_2^*(A, W_1^*(A), G^*(A)), D_{cls}^{\mathrm(val)}) \\
        \text{s.t.}\ & W_2^*(A, W_1^*(A), G^*(A)) = \min_{W_2} L(A, W_2, D_{cls}^{\mathrm(tr)})\ + \\
        & \hspace{1em} \lambda \sum_{c=1}^{C} l_c(A, W_1^*(A), D_{cls}^{\mathrm(val)}) \sum_{m=1}^{M} L(A, W_2, \hat{x_{c,m}}, c) \\
        & W_1^*(A) = \min_{W_1} L(A, W_1, D_{cls}^{\mathrm(tr)}) \\
        & G^*(A), H^*(A) = \min_{G} \max_{H} L(G, H, A, D_{cig})
    \end{aligned}
\end{equation}

\subsection{Optimization Algorithm}

In this section, we develop an efficient algorithm to solve three-level LFM-CW formulation strategy. Following \cite{Liu2019DARTSDA}, we approximate $W_1^*(A)$ using one-step gradient descent of $W_1$ w.r.t. $L(A, W_1, D_{cls}^{\mathrm(tr)})$ as:

\begin{equation}
\label{eq:eq_6}
    W_1^\prime = W_1 - \xi_{W_1} \nabla_{W_1} L(A, W_1, D_{cls}^{\mathrm(tr)})
\end{equation}

where $\xi_{W_1}$ is a learning rate for $W_1$. Similarly, we approximate $G^*(A)$ and $H^*(A)$ using one-step gradient descent update of $G$ and $H$ w.r.t. $L(G, H, A, D_{cig})$ as:

\begin{equation}
\label{eq:eq_7}
    G^\prime = G - \xi_{G} \nabla_{G} L(G, H, A, D_{cig})
\end{equation}

\begin{equation}
\label{eq:eq_8}
    H^\prime = H + \xi_{H} \nabla_{H} L(G, H, A, D_{cig})
\end{equation}

where $\xi_G$ and $\xi_H$ are learning rates for $G$ and $H$ respectively. We plug approximation $W_1^\prime$ of $W_1^*(A)$ in Eq.(\ref{eq:eq_3}) to obtain an approximated objective $O_{W_2}$:

\begin{multline}
\label{eq:eq_9}
    O_{W_2} = \min_{W_2} L(A, W_2, D_{cls}^{\mathrm(tr)}) + \\ \lambda \sum_{c=1}^{C} l_c(A, W_1^\prime, D_{cls}^{\mathrm(val)}) \sum_{m=1}^{M} L(A, W_2, \hat{x_{c,m}}, c)
\end{multline}

Next, we approximate $W_2^*(A, W_1^\prime, G^\prime)$ using one-step gradient descent update of $W_2$ w.r.t. $O_{W_2}$:

\begin{multline}
\label{eq:eq_10}
   W_2^\prime = W_2 - \xi_{W_2} \{\nabla_{W_2} L(A, W_2, D_{cls}^{\mathrm(tr)})\ + \\
   \lambda \sum_{c=1}^{C} l_c(A, W_1^\prime, D_{cls}^{\mathrm(val)}) \sum_{m=1}^{M} \nabla_{W_2} L(A, W_2, \hat{x_{c,m}}, c)\}
\end{multline}

where $\xi_{W_2}$ is a learning rate for $W_2$. Finally, we can update the architecture $A$ by calculating gradient of $L(A, W_2^\prime, D_{cls}^{\mathrm(val)})$ w.r.t. $A$ in the following way:

\begin{equation}
\label{eq:eq_11}
    A^\prime = A - \xi_{A} \nabla_{A} L(A, W_2^\prime, D_{cls}^{\mathrm(val)})
\end{equation}

where $\xi_{A}$ is a learning for $A$. Following chain rule, we can calculate $\nabla_{A} L(A, W_2^\prime, D_{cls}^{\mathrm(val)})$ in Eq.(\ref{eq:eq_11}) in the following way:

\begin{multline}
\label{eq:eq_12}
    \nabla_{A} L(A, W_2^\prime, D_{cls}^{\mathrm(val)}) = \frac{\partial L(A, W_2^\prime, D_{cls}^{\mathrm(val)})}{\partial A}\ + \\
    \hfill \frac{\partial W_2^\prime}{\partial A}\frac{\partial L(A, W_2^\prime, D_{cls}^{\mathrm(val)})}{\partial W_2^\prime}
\end{multline}

where

\begin{multline}
\label{eq:eq_13}
    \frac{\partial W_2^\prime}{\partial A}\frac{\partial L(A, W_2^\prime, D_{cls}^{\mathrm(val)})}{\partial W_2^\prime} = \frac{\partial W_2^\prime}{\partial A}\frac{\partial L(A, W_2^\prime, D_{cls}^{\mathrm(val)})}{\partial W_2^\prime}\ + \\
    \hspace{8em} \frac{\partial W_1^\prime}{\partial A}\frac{\partial W_2^\prime}{\partial W_1^\prime}\frac{\partial L(A, W_2^\prime, D_{cls}^{\mathrm(val)})}{\partial W_2^\prime}\ + \\
    \frac{\partial G^\prime}{\partial A}\frac{\partial W_2^\prime}{\partial G^\prime}\frac{\partial L(A, W_2^\prime, D_{cls}^{\mathrm(val)})}{\partial W_2^\prime}
\end{multline}

\begin{multline}
\label{eq:eq_14}
    \frac{\partial W_2^\prime}{\partial A} = -\xi_{W_2}\{\nabla_{A, W_2}^2 L(A, W_2, D_{cls}^{\mathrm(tr)}))\ + \\
    \lambda \sum_{i=1}^{C} \{\nabla_{A} l_c(A, W_1^\prime, D_{cls}^{\mathrm(val)}) \sum_{m=1}^{M} \nabla_{W_2} L(A, W_2, \hat{x_{c,m}}, c)\ + \\
    l_c(A, W_1^\prime, D_{cls}^{\mathrm(val)}) \sum_{m=1}^{M} \nabla_{A, W_2}^2 L(A, W_2, \hat{x_{c,m}}, c)\}\}
\end{multline}

\begin{multline}
\label{eq:eq_15}
    \frac{\partial W_2^\prime}{\partial W_1^\prime} = -\xi_{W_2} \{
    \lambda \sum_{i=1}^{C} \nabla_{W_1^\prime} l_c(A, W_1^\prime, D_{cls}^{\mathrm(val)}) \\ \sum_{m=1}^{M} \nabla_{W_2} L(A, W_2, \hat{x_{c,m}}, c)\}
\end{multline}

\begin{equation}
\label{eq:eq_16}
    \begin{aligned}
        \frac{\partial W_1^\prime}{\partial A} 
            &= \frac{\partial (W_1 - \xi_{W_1} \nabla_{W_1} L(A, W_1, D_{cls}^{\mathrm(tr)}))}{\partial A} \\
            &= -\xi_{W_1} \nabla_{A, W_1}^2 L(A, W_1, D_{cls}^{\mathrm(tr)}) \\
    \end{aligned}
\end{equation}

\begin{multline}
\label{eq:eq_17}
    \frac{\partial W_2^\prime}{\partial G^\prime} = -\xi_{W_2} \{ \lambda \sum_{i=1}^{C} l_c(A, W_1^\prime, D_{cls}^{\mathrm(val)}) \\ \sum_{m=1}^{M} \nabla_{G^\prime, W_2}^2 L(A, W_2, \hat{x_{c,m}}, c) \}
\end{multline}

\begin{equation}
\label{eq:eq_18}
    \begin{aligned}
        \frac{\partial G^\prime}{\partial A} 
            &= \frac{\partial (G - \xi_{G} \nabla_{G} L(G, H, A, D_{cig}))}{\partial A} \\
            &= -\xi_{G} \nabla_{A, G}^2 L(G, H, A, D_{cig}))
    \end{aligned}
\end{equation}

The overall algorithm for solving LFM-CW formulation is in Algorithm \ref{algo:f2_algo}

\begin{algorithm}[h]
  \caption{Optimization algorithm for LFM-CW formulation}
  \label{algo:f2_algo}
    \begin{algorithmic}
        \WHILE{not converged}
            \STATE Update the first set of weights $W_1$ using Eq.(\ref{eq:eq_6})
            \STATE Update the generator $G$ and discriminator $H$ using Eq.(\ref{eq:eq_7}) and Eq.(\ref{eq:eq_8})
            \STATE Update the second set of weights $W_2$ using Eq.(\ref{eq:eq_10})
            \STATE Update the architecture $A$ using Eq.(\ref{eq:eq_11})
        \ENDWHILE
    \end{algorithmic}
\end{algorithm}

\section{Experiments}
\label{sec:experiments}

In this section, we apply our proposed LFM-CW strategy to image classification tasks. We incorporate it into differential architecture search approaches such as DARTS, P-DARTS, and PC-DARTS and perform experiments on CIFAR and ImageNet datasets. Following \cite{Liu2019DARTSDA}, each experiment is composed of two phases: architecture search and evaluation. In the search phase, we find out an optimal cell by minimizing the validation loss. In the evaluation phase, multiple copies of an optimal searched cell are stacked and composed into a larger network, which we train from scratch and report their performance on the test set.

\subsection{Datasets}
\label{subsec:datasets}

We conduct experiments on three image classification datasets: CIFAR-10 \cite{Krizhevsky2009LearningML}, CIFAR-100 \cite{Krizhevsky2009LearningML}, and ImageNet \cite{Deng2009ImageNetAL}. Both CIFAR-10 and CIFAR-100 datasets contain $50K$ training images and $10K$ testing images, with 10 and 100 classes (the number of images in each class is equal), respectively. For each of them, we split the original $50K$ training images into a new $25K$ training set and $25K$ validation set. ImageNet dataset contains $1.2M$ training images and $50K$ validation images, with 1000 classes. The validation set is used as a test set for architecture evaluation. Architecture search on the $1.2M$ training images is computationally too expensive. To overcome this issue, following \cite{Xu2020PCDARTSPC}, we randomly sample $10\%$ and $2.5\%$ images from the $1.2M$ training images to form a new training set and validation set, respectively. We perform the architecture search on the newly obtained subset and train the large architecture composed of multiple candidate cells on the full set of $1.2M$ images during evaluation.

\begin{table*}[t]
    \caption{Test error on CIFAR datasets with different NAS algorithms. Results indicated with * are provided from Skillearn \cite{Xie2020SkillearnML}.}
    \label{tab:cifar}
    \vskip 0.15in
    \begin{center}
    \begin{small}
    \begin{tabular}{{p{5.1cm} p{1.6cm} p{1.7cm} p{1.6cm} p{1.6cm}  p{1.6cm}}}
    \toprule
    Methods & CIFAR-10 Error (\%) & CIFAR-100 Error (\%) & CIFAR-10 Param (M) & CIFAR-100 Param (M) & Search Cost (GPU-days) \\
    \midrule
    *DenseNet \cite{Huang2017DenselyCC} & 3.46 & 17.18 & 25.6 & 25.6 & - \\
    \midrule
    *PNAS \cite{Liu2018ProgressiveNA} & 3.41 $\pm$ 0.09 & 19.53 & 3.2 & 3.2 & 225 \\
    *ENAS \cite{Pham2018EfficientNA} & 2.89 & 19.43 & 4.6 & 4.6 & 0.5 \\
    *AmoebaNet \cite{Real2019RegularizedEF} & 2.55 $\pm$ 0.05 & 18.93 & 2.8 & 3.1 & 3150 \\
    *HierEvol \cite{Liu2018HierarchicalRF} & 3.75 & - & 15.7 & - & 300 \\
    *GDAS \cite{Dong2019SearchingFA} & 2.93 & 18.38 & 3.4 & 3.4 & 0.2 \\
    *DropNAS \cite{Hong2020DropNASGO} & 2.58 $\pm$ 0.14 & 16.39 & 4.1 & 4.4 & 0.7 \\
    *ProxylessNAS \cite{Cai2019ProxylessNASDN} & 2.08 & - & 5.7 & - & 4.0 \\
    *GTN\cite{Such2020GenerativeTN} & 2.92 $\pm$ 0.06 & - & 8.2 & - & 0.67 \\
    *BayesNAS \cite{Zhou2019BayesNASAB} & 2.81 $\pm$ 0.04 & - & 3.4 & - & 0.2 \\
    *MergeNAS \cite{Wang2020MergeNASMO} & 2.73 $\pm$ 0.02 & - & 2.9 & - & 0.2 \\
    *NoisyDARTS \cite{Chu2020NoisyDA} & 2.70 $\pm$ 0.23 & - & 3.3 & - & 0.4 \\
    *ASAP \cite{Noy2020ASAPAS} & 2.68 $\pm$ 0.11 & - & 2.5 & - & 0.2 \\
    *SDARTS \cite{Chen2020StabilizingDA} & 2.61 $\pm$ 0.02 & - & 3.3 & - & 1.3 \\
    *FairDARTS \cite{Chu2020FairDE} & 2.54 & - & 3.3 & - & 0.4 \\
    *DrNAS \cite{Chen2021DrNASDN} & 2.54 $\pm$ 0.03 & - & 4.0 & - & 0.4 \\
    *R-DARTS \cite{Zela2020UnderstandingAR} & 2.95 $\pm$ 0.21 & 18.01 $\pm$ 0.26 & - & - & 1.6 \\
    *DARTS$^{-}$ \cite{Chu2021DARTSRS} & 2.59 $\pm$ 0.08 & 17.51 $\pm$ 0.25 & 3.5 & 3.3 & 0.4 \\
    *DARTS$^{-}$ \cite{Chu2021DARTSRS} & 2.97 $\pm$ 0.04 & 18.97 $\pm$ 0.16 & 3.3 & 3.1 & 0.4 \\
    *DARTS$^{+}$ \cite{Liang2019DARTSID} & 2.83 $\pm$ 0.05 & - & 3.7 & - & 0.4 \\
    *DARTS$^{+}$ \cite{Liang2019DARTSID} & - & 17.11 $\pm$ 0.43 & - & 3.8 & 0.2 \\
    \midrule
    *DARTS-1st \cite{Liu2019DARTSDA} & 3.00 $\pm$ 0.14 & 20.52 $\pm$ 0.31 & 3.3 & 3.5 & 0.4 \\
    *DARTS-2nd \cite{Liu2019DARTSDA} & 2.76 $\pm$ 0.09 & 20.58 $\pm$ 0.44 & 3.3& 3.5 & 1.5 \\
    $\;\;$LFM-CW-BigGAN-DARTS-1st (ours) & 2.55 $\pm$ 0.06 & 16.59 $\pm$ 0.08 & 2.5 & 3.2 & 1.5 \\
    $\;\;$LFM-CW-BigGAN-DARTS-2nd (ours) & \textbf{2.54 $\pm$ 0.10} & \textbf{16.42 $\pm$ 0.12} & 3.6 & 3.4 & 2.8 \\
    \midrule
    *PC-DARTS \cite{Xu2020PCDARTSPC} & 2.57 $\pm$ 0.07 & 17.96 $\pm$ 0.15 & 3.6 & 3.9 & 0.1 \\
    $\;\;$LFM-CW-BigGAN-PCDARTS (ours) & \textbf{2.48 $\pm$ 0.06} & \textbf{16.4 $\pm$ 0.16} & 3.3 & 3.0 & 0.7 \\
    \midrule
    *P-DARTS \cite{Chen2019ProgressiveDA} & 2.50 & 17.49 & 3.4 & 3.6 & 0.3 \\
    $\;\;$LFM-CW-BigGAN-PDARTS (ours) & \textbf{2.53 $\pm$ 0.05} & \textbf{15.96 $\pm$ 0.17} & 3.2 & 3.1 & 0.9 \\
    \bottomrule
    \end{tabular}
    \end{small}
    \end{center}
    \vskip -0.1in
\end{table*}

\subsection{Experimental Settings}
\label{subsec:exp_settings}

Our approach is agnostic that can be coupled with any existing differential search method. Specifically, we apply our approach to the following NAS methods: 1) DARTS \cite{Liu2019DARTSDA} 2) P-DARTS \cite{Chen2019ProgressiveDA}, and 3) PC-DARTS \cite{Xu2020PCDARTSPC}, where the search space is composed of building blocks such as $3\times3$ and $5\times5$ (dilated) separable convolutions, $3\times3$ max pooling, $3\times3$ average pooling, zero, and identity. We use BigGAN \cite{Brock2019LargeSG} for the CIG in the proposed LFM-CW formulation. The hyperparameter $\lambda$ is set to 1. We generate synthetic images of the same target labels as the training images in a mini-batch for class-wise weighted loss described in Eq.(\ref{eq:eq_3}) (as displayed in Figure \ref{fig:cifar_samples} and Figure \ref{fig:imagenet_samples} in the appendix \ref{app:visualization}). We report the mean and standard deviation of classification errors obtained from 10 runs with different random seeds.

During architecture search for both CIFAR-10 and CIFAR-100, each architecture consists of a stack of 8 cells, and each cell consists of 7 nodes, with the initial channel number set to 16. SGD (Stochastic Gradient Descent) optimizer was used to optimize network weights, with an initial learning rate of 0.025, a weight decay of 3e-4, and a momentum of 0.9. Cosine decay scheduler \cite{Loshchilov2017SGDRSG} was used for scheduling the learning rate. Adam optimizer \cite{Kingma2015AdamAM} was used to optimize the architecture, with a learning rate of 3e-4 and a weight decay of 1e-3. The generator and discriminator in the CIG are optimized using Adam optimizer, with a learning rate of 2e-4. The search was performed for 50 epochs in DARTS and PC-DARTS variants and 25 epochs in the P-DARTS variant, with a batch size of 64, respectively.

During the architecture evaluation for CIFAR-10 and CIFAR-100, a larger network is composed by stacking 20 copies of the searched cell, with the initial channel number set to 36. The network is trained for 600 epochs with a batch size of 96 on a single Tesla V100. SGD optimizer is used for network weights training, with an initial learning rate of 0.025, a cosine decay scheduler, a weight decay of 3e-4, and a momentum of 0.9. For ImageNet, we evaluate two types of architectures: 1) those searched on CIFAR-10 and CIFAR-100; 2) those searched on a subset of ImageNet. In either type, we stack 14 copies of an optimally searched cell into a larger network, with the initial channel number set to 48. The network is trained for 250 epochs, with an initial learning rate of 0.5, a weight decay of 3e-5, a momentum of 0.9, and a batch size of 1024 on eight Tesla V100s. The rest of the hyperparameters for search and evaluation phase follow the same settings as those in DARTS, P-DARTS, PC-DARTS (detailed list mentioned in the appendix \ref{app:params_settings}).

\subsection{Results}

\begin{table*}[t]
    \caption{Test error on ImageNet dataset with various NAS algorithms. Results highlighted with * are obtained from Skillearn \cite{Xie2020SkillearnML}.}
    \label{tab:imagenet}
    \vskip 0.15in
    \begin{center}
    \begin{small}
    \begin{tabular}{{p{6.6cm} p{1.25cm} p{1.25cm} p{1.4cm} p{1.6cm}}}
    \toprule
    Methods & Top-1\ \ \ \ Error (\%) & Top-5\ \ \ \ Error (\%) & Param (M) & Search Cost (GPU-days) \\
    \midrule
    *Inception-v1 \cite{Szegedy2015GoingDW} & 30.2 & 10.1 & 6.6 & - \\
    *MobileNet \cite{Howard2017MobileNetsEC} & 29.4 & 10.5 & 4.2 & - \\
    *ShuffleNet 2$\times$ (v1) \cite{Zhang2018ShuffleNetAE} & 26.4 & 10.2 & 5.4 & - \\
    *ShuffleNet 2$\times$ (v2) \cite{Ma2018ShuffleNetVP} & 25.1 & 7.6 & 7.4 & - \\
    \midrule
    *NASNet-A \cite{Zoph2018LearningTA} & 26.0 & 8.4 & 5.3 & 1800 \\
    *PNAS \cite{Liu2018ProgressiveNA} & 25.8 & 8.1 & 5.1 & 225 \\
    *MnasNet-92 \cite{Tan2019MnasNetPN} & 25.2 & 8.0 & 4.4 & 1667 \\
    *AmoebaNet-C \cite{Real2019RegularizedEF} & 24.3 & 7.6 & 6.4 & 3150 \\
    *SNAS-CIFAR10 \cite{Xie2019SNASSN} & 27.3 & 9.2 & 4.3 & 1.5 \\
    *BayesNAS-CIFAR10 \cite{Zhou2019BayesNASAB} & 26.5 & 8.9 & 3.9 & 0.2 \\
    *PARSEC-CIFAR10 \cite{Casale2019ProbabilisticNA} & 26.0 & 8.4 & 5.6 & 1.0 \\
    *GDAS-CIFAR10 \cite{Dong2019SearchingFA} & 26.0 & 8.5 & 5.3 & 0.2 \\
    *DSNAS-ImageNet \cite{Hu2020DSNASDN} & 25.7 & 8.1 & - & - \\
    *SDARTS-ADV-CIFAR10 \cite{Chen2020StabilizingDA} & 25.2 & 7.8 & 5.4 & 1.3 \\
    *ProxylessNAS-ImageNet \cite{Cai2019ProxylessNASDN} & 24.9 & 7.5 & 7.1 & 8.3 \\
    *FairDARTS-CIFAR10 \cite{Chu2020FairDE} & 24.9 & 7.5 & 4.8 & 0.4 \\
    *FairDARTS-ImageNet \cite{Chu2020FairDE} & 24.4 & 7.4 & 4.3 & 3.0 \\
    *DrNAS-ImageNet \cite{Chen2021DrNASDN} & 24.2 & 7.3 & 5.2 & 3.9 \\
    *DARTS$^{+}$-ImageNet \cite{Liang2019DARTSID}& 23.9 & 7.4 & 5.1 & 6.8 \\
    *DARTS$^{-}$-ImageNet \cite{Chu2021DARTSRS} & 23.8 & 7.0 & 4.9 & 4.5 \\
    *DARTS$^{+}$-CIFAR100 \cite{Liang2019DARTSID} & 23.7 & 7.2 & 5.1 & 0.2 \\
    \midrule
    *DARTS-2nd-CIFAR10 \cite{Liu2019DARTSDA} & 26.7 & 8.7 & 4.7 & 1.5 \\
    ${}^{\dag}$DARTS-1st-CIFAR10 \cite{Liu2019DARTSDA} & 26.1 & 8.3 & 4.5 & 0.4 \\
    $\;\;$LFM-CW-BigGAN-DARTS-1st-CIFAR100 (ours) & 25.5 & 8.0 & 4.6 & 1.5 \\
    $\;\;$LFM-CW-BigGAN-DARTS-2nd-CIFAR10 (ours) & 25.3 & 7.9 & 5.1 & 2.8 \\
    $\;\;$LFM-CW-BigGAN-DARTS-2nd-CIFAR100 (ours) & \textbf{25.3} & \textbf{7.6} & 4.5 & 2.8 \\
    \midrule
    *PDARTS (CIFAR10) \cite{Chen2019ProgressiveDA} & 24.4 & 7.4 & 4.9 & 0.3 \\
    $\;\;$LFM-CW-BigGAN-PDARTS-CIFAR10 (ours) & \textbf{23.6} & \textbf{7.0} & 4.8 & 0.9 \\
    \midrule
    *PCDARTS-CIFAR10 \cite{Xu2020PCDARTSPC} & 25.1 & 7.8 & 5.3 & 0.1 \\
    *PCDARTS-ImageNet \cite{Xu2020PCDARTSPC} & 24.2 & 7.3 & 5.3 & 3.8 \\
      $\;\;$LFM-CW-BigGAN-PCDARTS-ImageNet (ours) & \textbf{22.6} & \textbf{6.2} & 5.1 & 4.2 \\
    \bottomrule
    \end{tabular}
    \end{small}
    \end{center}
    \vskip -0.1in
\end{table*}

\begin{figure*}[htb]
    \vskip 0.2in
    \begin{center}
    \subfloat[DARTS-1st]{\includegraphics[width=0.25\linewidth]{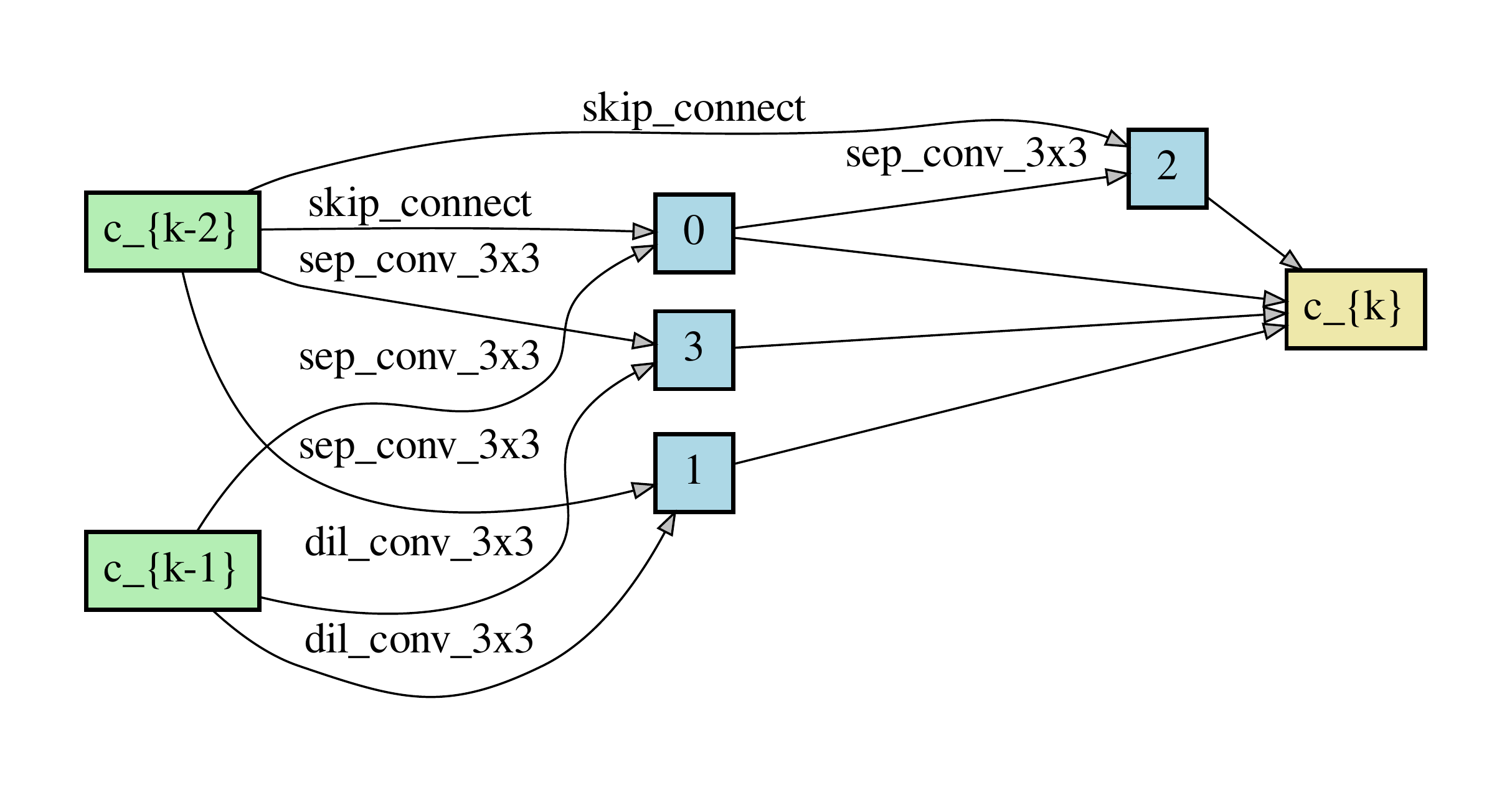}}
    \subfloat[DARTS-2nd]{\includegraphics[width=0.25\linewidth]{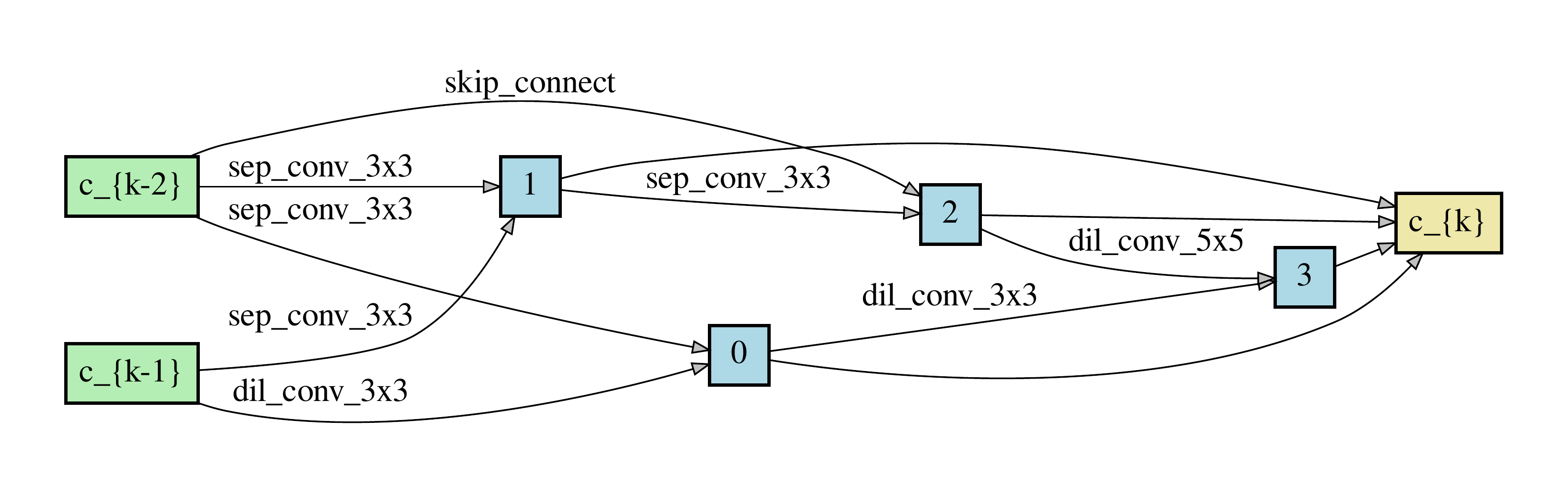}}
    \subfloat[P-DARTS]{\includegraphics[width=0.25\linewidth]{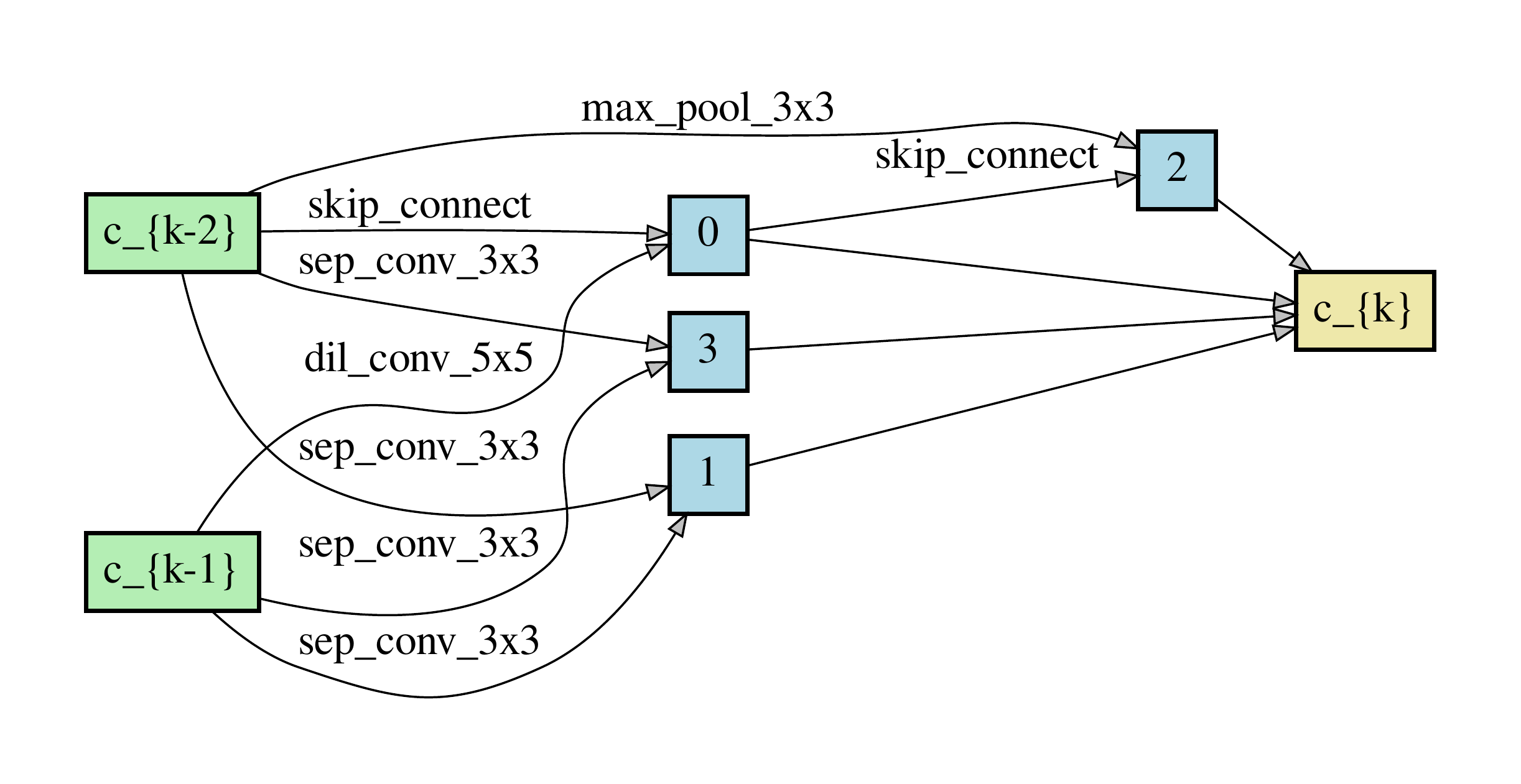}}
    \subfloat[PC-DARTS]{\includegraphics[width=0.25\linewidth]{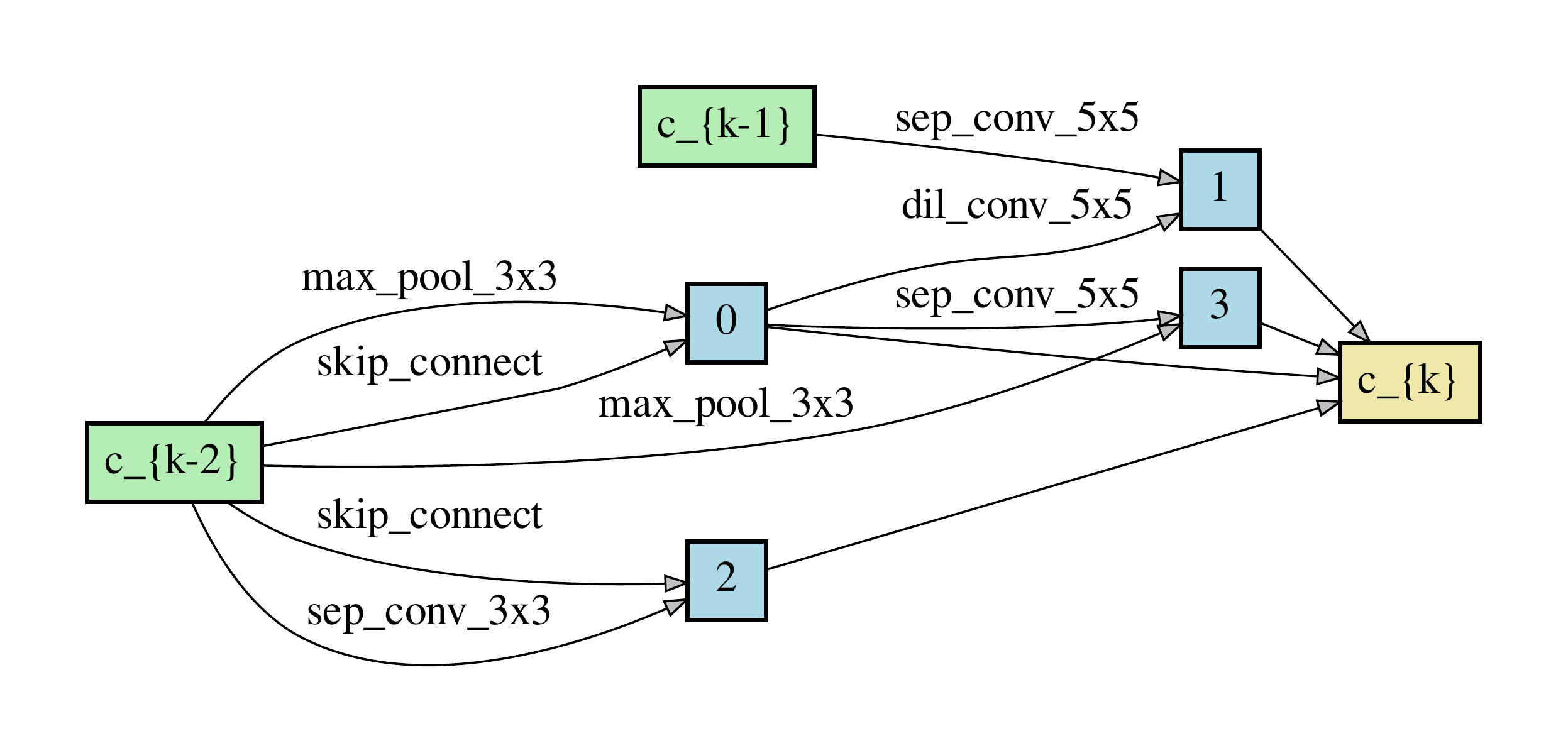}}
    \caption{Discovered normal cells on CIFAR-100 dataset during architecture search using our approach with DARTS-1st, DARTS-2nd, P-DARTS, and PC-DARTS search space.}
    \label{fig:arch_normal}
    \end{center}
    \vskip -0.2in
\end{figure*}

Table \ref{tab:cifar} shows the classification error (\%) of different NAS methods on the CIFAR-10 and CIFAR-100 test sets, including the number of parameters (millions), and search cost (GPU days). We can make the following observations from the table. Our proposed LFM-CW framework significantly reduces the errors on the test set when compared to existing differential NAS methods such as DARTS, P-DARTS, and PC-DARTS. This highlights the efficacy of our method for searching for better architectures on a target task. For example, applying LFM-CW to DARTS on CIFAR-100 reduces the error by approximately 4\% for first-order and second-order approximation. As another example, our method reduces the error by approximately 1.5\% for P-DARTS and PC-DARTS on CIFAR-100. Similarly, these observations can be further extended to the architecture search on CIFAR-10, where our method achieves superior performance compared to the baselines. Overall, our method achieves superior performance on CIFAR-100 than that on CIFAR-10. This is likely because CIFAR-10 is a relatively simple dataset to classify, containing only 10 classes, leaving little scope for improvement. CIFAR-100 is a more challenging dataset for classification due to 100 classes and can better differentiate the capabilities of different methods. Figure \ref{fig:arch_normal} illustrates the visualization of searched normal cells on the CIFAR-100 dataset using DARTS, P-DARTS, and PC-DARTS search space.

Table \ref{tab:imagenet} shows the classification error (\%) of different NAS methods on the test set of ImageNet, the number of model parameters (millions), and search cost (GPU days). Our method outperforms PC-DARTS on ImageNet and achieves the lowest error (22.6\% top-1 error and 6.2\% top-5 error) among all the methods. For other experiments, the architectures are searched on CIFAR-10 and CIFAR-100 and then evaluated on ImageNet. As observed from the results, the architectures searched by our proposed technique are better than the corresponding baselines. For example, LFM-CW-BigGAN-PDARTS-CIFAR10 achieves a lower error than PDARTS-CIFAR10. Similar observations can be made for the DARTS search space. This further demonstrates the effectiveness of our method.

\subsection{Abalation Study}

In this section, we perform the following ablation studies on LFM-CW formulation to understand the significance and effectiveness of different modules. For the ablation studies, we use the training and validation set for search and evaluation, same as mentioned in section \ref{subsec:datasets}.

\begin{figure}[ht]
    \vskip 0.2in
    \begin{center}
    \includegraphics[width=0.495\columnwidth]{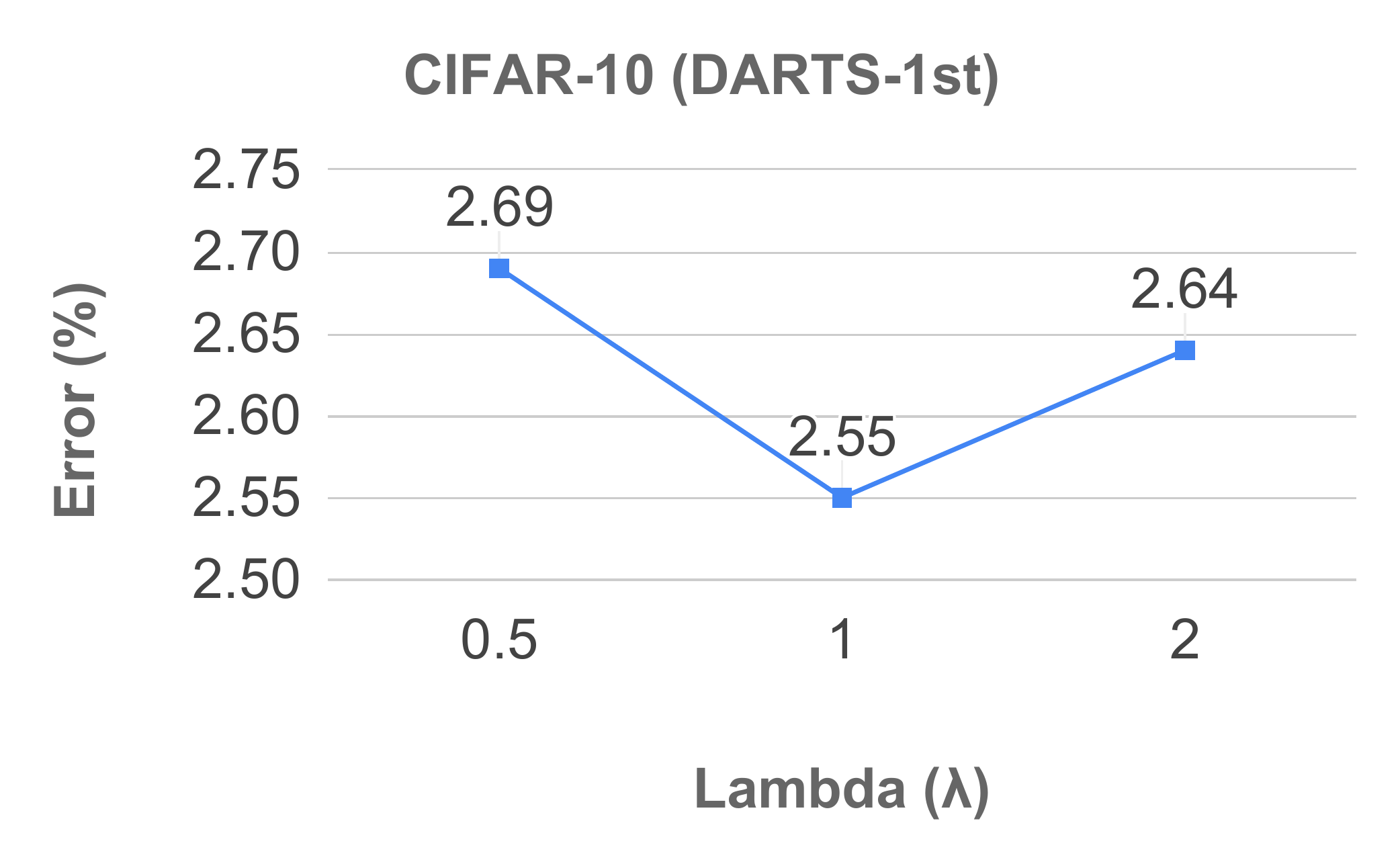}
    \includegraphics[width=0.495\columnwidth]{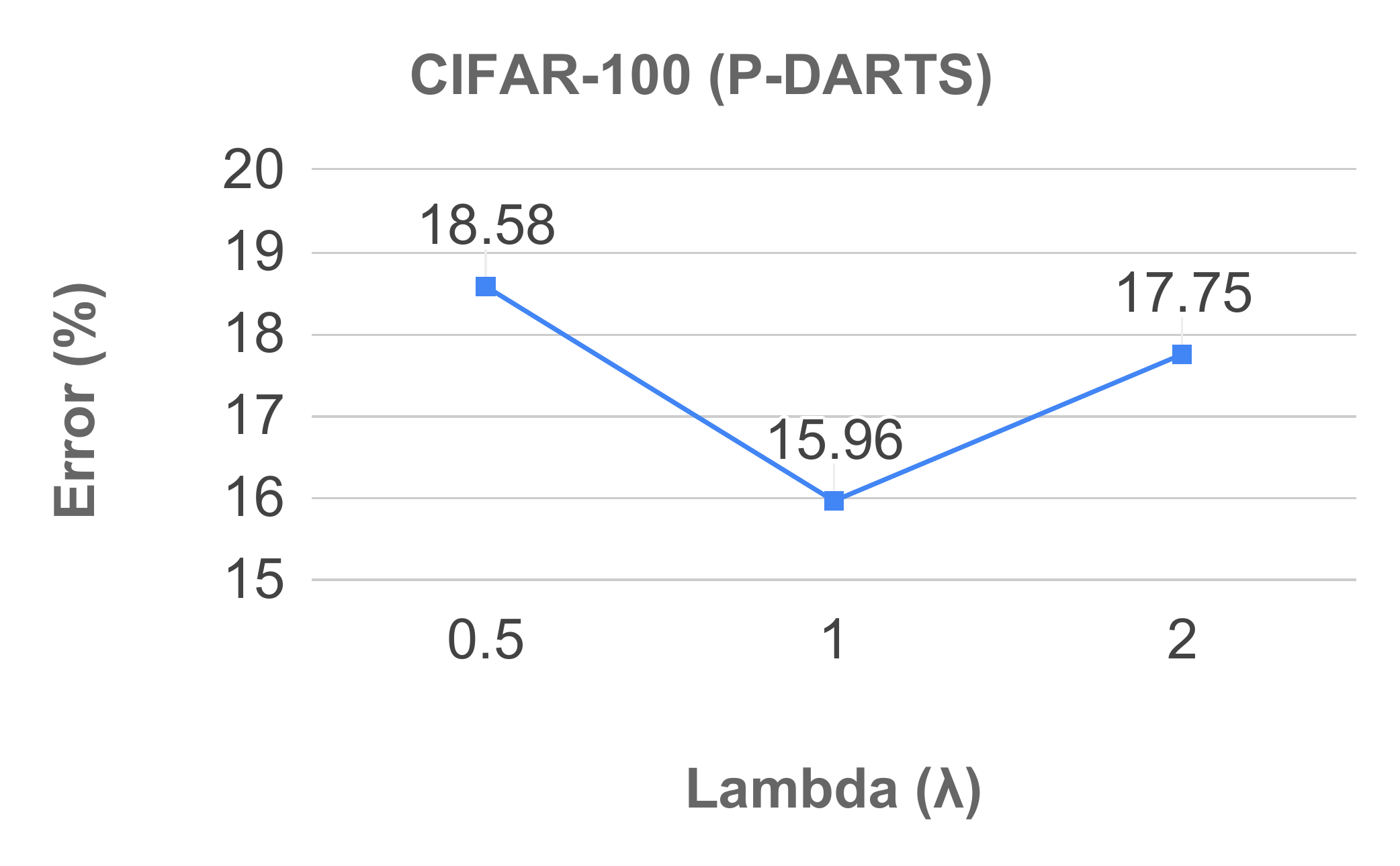}
    \caption{How errors change as $\lambda$ increases.}
    \label{fig:abl_study}
    \end{center}
    \vskip -0.2in
\end{figure}

\begin{itemize}[leftmargin=*]
    \item \textbf{Ablation Setting 1}. In this setting, we investigate how the trade-off parameter $\lambda$ in Eq.(\ref{eq:eq_5}) affects the classification error. We apply LFM-CW formulation to DARTS-1st for CIFAR-10 and P-DARTS for CIFAR-100. Figure \ref{fig:abl_study} shows how classification error on the test sets of CIFAR-10 and CIFAR-100 vary as $\lambda$ increases. We can observe that increasing $\lambda$ from 0.5 to 1 minimizes the error on both CIFAR-10 and CIFAR-100. However, further increasing $\lambda$ increases the error. We conjecture that increasing $\lambda$ might coerce the network to emphasize training signals from synthetic images rather than actual training images, resulting in inferior performance since generated synthetic images are of poor quality and noisy during initial learning.
    
    \item \textbf{Ablation Setting 2}. In this setting, the classification model with the second set of weights $W_2$ updates its weights by minimizing the loss on adversarially generated images only, without considering the loss on training set images. The corresponding formulation is following:
    \begin{equation}\label{eq:eq_19}
        \begin{aligned}
            \min_{A}\ & L(W_2^*(A, W_1^*(A), G^*(A)), D_{cls}^{\mathrm(val)}) \\
            \text{s.t.}\ & W_2^*(A, W_1^*(A), G^*(A)) = \\
            &\lambda \sum_{c=1}^{C} l_c(A, W_1^*(A), D_{cls}^{\mathrm(val)}) \sum_{m=1}^{M} L(A, W_2, \hat{x_{c,m}}, c) \\
            & W_1^*(A) = \min_{W_1} L(A, W_1, D_{cls}^{\mathrm(tr)}) \\
            & G^*(A), H^*(A) = \min_{G, H} L(G, H, A, D_{cig})
        \end{aligned}
    \end{equation}
    
    \begin{table}[t]
        \caption{Results for ablation setting 2. ``Synthetic + training data" denotes the proposed LFM-CW formulation without modification whereas the ``Synthetic data" denotes the formulation in Eq.(\ref{eq:eq_19}) with only synthetic data used for updating $W_2$ in the second learning stage.}
        \label{tab:abl2}
        \vskip 0.15in
        \begin{center}
        \begin{small}
        \begin{tabular}{p{4.5cm} p{1.65cm}}
            \toprule
            Method & Error (\%) \\
            \midrule
            Synthetic data only (DARTS-1st CIFAR-10) & 3.62 $\pm$ 0.08 \\
            Synthetic + training data (DARTS-1st CIFAR-10) & \textbf{2.55 $\pm$ 0.06} \\
            \midrule
            Synthetic data only (P-DARTS CIFAR-100) & 22.22 $\pm$ 0.25 \\
            Synthetic + training data (P-DARTS CIFAR-100) & \textbf{15.96 $\pm$ 0.17} \\
            \bottomrule
            \end{tabular}
        \end{small}
        \end{center}
        \vskip -0.1in
    \end{table}

    During this study, our proposed LFM-CW framework is applied to DARTS-1st on CIFAR-10 and P-DARTS on CIFAR-100. Table \ref{tab:abl2} shows the classification error on CIFAR-10 and CIFAR-100 test sets for the ablation setting 2. On both datasets, we can see that combining losses from both the synthetic data and training data to train the classification network performs better than just using loss from the synthetic data in the second stage. Loss from training data acts as a way of regularization, which does not allow the classification network to diverge from the true distribution while also leveraging noise from the synthetic data to help the classification model learn better.
    
    \item \textbf{Ablation Setting 3}. In this setting, we investigate how the classification error changes with the different conditional image generators. The experiment uses the following image generators in the first learning stage: DCGAN \cite{Radford2016UnsupervisedRL}, WGANGP \cite{Gulrajani2017ImprovedTO}, and BigGAN \cite{Brock2019LargeSG}. On both datasets, the LFM-CW framework is applied to DARTS-1st. Table \ref{tab:abl3} demonstrates that the classification error reduces with the use of a sophisticated deep conditional image generator on both datasets. The intuitive reason is that: better image generator results in a better classification model in the second stage, which further results in better architecture, improving the overall learning.
    
    \begin{table}[t]
        \caption{Results for ablation setting 3. Test error of searched architectures with different conditional image generators in first stage on CIFAR datasets.}
        \label{tab:abl3}
        \vskip 0.15in
        \begin{center}
        \begin{small}
        \begin{tabular}{p{4.65cm} p{1.65cm}}
            \toprule
            Method & Error (\%) \\
            \midrule
            DCGAN-DARTS-1st (CIFAR-10) & 2.92 $\pm$ 0.10 \\
            WGANGP-DARTS-1st (CIFAR-10) & 2.81 $\pm$ 0.15 \\
            BigGAN-DARTS-1st (CIFAR-10) & \textbf{2.55 $\pm$ 0.06} \\
            \midrule
            DCGAN-DARTS-1st (CIFAR-100) & 22.09 $\pm$ 0.42 \\
            WGANGP-DARTS-1st (CIFAR-100) & 20.77 $\pm$ 0.19 \\
            BigGAN-DARTS-1st (CIFAR-100) & \textbf{16.59 $\pm$ 0.08} \\
            \bottomrule
        \end{tabular}
        \end{small}
        \end{center}
        \vskip -0.1in
    \end{table}
\end{itemize}

\section{Conclusion}
\label{sec:conclusion}

In this paper, we present a simple yet effective approach --- learning from mistakes using class weighting (LFM-CW), motivated by error-driven learning to search architecture for deep convolutional networks. The core idea is to enforce the neural network to pay greater attention to the classes, i.e., distributions for which it is making errors to avoid making similar errors in the future. Our proposed LFM-CW framework involves learning the class weights by minimizing the validation loss of the model and re-training the model with the synthetic data weighted by class-wise performance and real data. Experimental results on different image classification benchmarks strongly demonstrate the effectiveness of our proposed method. 

An interesting future direction is to extend the current work for cross-dataset generalization similar to meta-learning by performing architecture search over a labeled dataset and extensively evaluating using supervised contrastive learning over the other unlabeled datasets in an interleaving manner. This would aid in developing effective systems in healthcare, where the human-annotated data is limited compared to unlabeled data. Another challenge is to reduce the computational resources required for our proposed strategy during the architecture search by using parameter sharing across models in different learning stages.


\bibliography{main}

\begin{thebibliography}{61}
\providecommand{\natexlab}[1]{#1}
\providecommand{\url}[1]{\texttt{#1}}
\expandafter\ifx\csname urlstyle\endcsname\relax
  \providecommand{\doi}[1]{doi: #1}\else
  \providecommand{\doi}{doi: \begingroup \urlstyle{rm}\Url}\fi

\bibitem[Axelrod et~al.(2011)Axelrod, He, and Gao]{Axelrod2011DomainAV}
Axelrod, A., He, X., and Gao, J.
\newblock Domain adaptation via pseudo in-domain data selection.
\newblock In \emph{EMNLP}, 2011.

\bibitem[Baker et~al.(2017)Baker, Gupta, Naik, and
  Raskar]{Baker2017DesigningNN}
Baker, B., Gupta, O., Naik, N., and Raskar, R.
\newblock Designing neural network architectures using reinforcement learning.
\newblock In \emph{ICLR}, 2017.

\bibitem[Brock et~al.(2019)Brock, Donahue, and Simonyan]{Brock2019LargeSG}
Brock, A., Donahue, J., and Simonyan, K.
\newblock Large scale gan training for high fidelity natural image synthesis.
\newblock In \emph{ICLR}, 2019.

\bibitem[Cai et~al.(2019)Cai, Zhu, and Han]{Cai2019ProxylessNASDN}
Cai, H., Zhu, L., and Han, S.
\newblock Proxylessnas: Direct neural architecture search on target task and
  hardware.
\newblock In \emph{ICLR}, 2019.

\bibitem[Casale et~al.(2019)Casale, Gordon, and
  Fusi]{Casale2019ProbabilisticNA}
Casale, F.~P., Gordon, J., and Fusi, N.
\newblock Probabilistic neural architecture search.
\newblock \emph{CoRR}, abs/1902.05116, 2019.

\bibitem[Chen \& Hsieh(2020)Chen and Hsieh]{Chen2020StabilizingDA}
Chen, X. and Hsieh, C.-J.
\newblock Stabilizing differentiable architecture search via perturbation-based
  regularization.
\newblock In \emph{ICML}, 2020.

\bibitem[Chen et~al.(2019)Chen, Xie, Wu, and Tian]{Chen2019ProgressiveDA}
Chen, X., Xie, L., Wu, J., and Tian, Q.
\newblock Progressive differentiable architecture search: Bridging the depth
  gap between search and evaluation.
\newblock pp.\  1294--1303, 2019.

\bibitem[Chen et~al.(2021)Chen, Wang, Cheng, Tang, and Hsieh]{Chen2021DrNASDN}
Chen, X., Wang, R., Cheng, M., Tang, X., and Hsieh, C.-J.
\newblock Drnas: Dirichlet neural architecture search.
\newblock In \emph{ICLR}, 2021.

\bibitem[Chrysos et~al.(2019)Chrysos, Kossaifi, and
  Zafeiriou]{Chrysos2019RobustCG}
Chrysos, G.~G., Kossaifi, J., and Zafeiriou, S.
\newblock Robust conditional generative adversarial networks.
\newblock In \emph{ICLR}, 2019.

\bibitem[Chu et~al.(2020{\natexlab{a}})Chu, Zhang, and Li]{Chu2020NoisyDA}
Chu, X., Zhang, B., and Li, X.
\newblock Noisy differentiable architecture search.
\newblock \emph{CoRR}, abs/2005.03566, 2020{\natexlab{a}}.

\bibitem[Chu et~al.(2020{\natexlab{b}})Chu, Zhou, Zhang, and Li]{Chu2020FairDE}
Chu, X., Zhou, T., Zhang, B., and Li, J.
\newblock Fair {DARTS:} eliminating unfair advantages in differentiable
  architecture search.
\newblock In \emph{ECCV}, 2020{\natexlab{b}}.

\bibitem[Chu et~al.(2021)Chu, Wang, Zhang, Lu, Wei, and Yan]{Chu2021DARTSRS}
Chu, X., Wang, X., Zhang, B., Lu, S., Wei, X., and Yan, J.
\newblock {DARTS-:} robustly stepping out of performance collapse without
  indicators.
\newblock In \emph{ICLR}, 2021.

\bibitem[Cui et~al.(2019)Cui, Jia, Lin, Song, and
  Belongie]{Cui2019ClassBalancedLB}
Cui, Y., Jia, M., Lin, T.-Y., Song, Y., and Belongie, S.~J.
\newblock Class-balanced loss based on effective number of samples.
\newblock \emph{2019 IEEE/CVF Conference on Computer Vision and Pattern
  Recognition (CVPR)}, pp.\  9260--9269, 2019.

\bibitem[Deng et~al.(2009)Deng, Dong, Socher, Li, Li, and
  Fei-Fei]{Deng2009ImageNetAL}
Deng, J., Dong, W., Socher, R., Li, L.-J., Li, K., and Fei-Fei, L.
\newblock Imagenet: A large-scale hierarchical image database.
\newblock In \emph{CVPR}, 2009.

\bibitem[Dong \& Yang(2019)Dong and Yang]{Dong2019SearchingFA}
Dong, X. and Yang, Y.
\newblock Searching for a robust neural architecture in four gpu hours.
\newblock \emph{2019 IEEE/CVF Conference on Computer Vision and Pattern
  Recognition (CVPR)}, pp.\  1761--1770, 2019.

\bibitem[Foster et~al.(2010)Foster, Goutte, and
  Kuhn]{Foster2010DiscriminativeIW}
Foster, G.~F., Goutte, C., and Kuhn, R.
\newblock Discriminative instance weighting for domain adaptation in
  statistical machine translation.
\newblock In \emph{EMNLP}, 2010.

\bibitem[Goodfellow et~al.(2014)Goodfellow, Pouget-Abadie, Mirza, Xu,
  Warde-Farley, Ozair, Courville, and Bengio]{Goodfellow2014GenerativeAN}
Goodfellow, I.~J., Pouget-Abadie, J., Mirza, M., Xu, B., Warde-Farley, D.,
  Ozair, S., Courville, A.~C., and Bengio, Y.
\newblock Generative adversarial nets.
\newblock In \emph{NIPS}, 2014.

\bibitem[Gulrajani et~al.(2017)Gulrajani, Ahmed, Arjovsky, Dumoulin, and
  Courville]{Gulrajani2017ImprovedTO}
Gulrajani, I., Ahmed, F., Arjovsky, M., Dumoulin, V., and Courville, A.~C.
\newblock Improved training of wasserstein gans.
\newblock In \emph{NIPS}, 2017.

\bibitem[He et~al.(2016)He, Zhang, Ren, and Sun]{He2016DeepRL}
He, K., Zhang, X., Ren, S., and Sun, J.
\newblock Deep residual learning for image recognition.
\newblock \emph{2016 IEEE Conference on Computer Vision and Pattern Recognition
  (CVPR)}, pp.\  770--778, 2016.

\bibitem[Hong et~al.(2020)Hong, Li, Zhang, Tang, Wang, Li, and
  Yu]{Hong2020DropNASGO}
Hong, W., Li, G., Zhang, W., Tang, R., Wang, Y., Li, Z., and Yu, Y.
\newblock Dropnas: Grouped operation dropout for differentiable architecture
  search.
\newblock In \emph{IJCAI}, 2020.

\bibitem[Howard et~al.(2017)Howard, Zhu, Chen, Kalenichenko, Wang, Weyand,
  Andreetto, and Adam]{Howard2017MobileNetsEC}
Howard, A.~G., Zhu, M., Chen, B., Kalenichenko, D., Wang, W., Weyand, T.,
  Andreetto, M., and Adam, H.
\newblock Mobilenets: Efficient convolutional neural networks for mobile vision
  applications.
\newblock \emph{CoRR}, abs/1704.04861, 2017.

\bibitem[Hu et~al.(2020)Hu, Xie, Zheng, Liu, Shi, Liu, and Lin]{Hu2020DSNASDN}
Hu, S.-Y., Xie, S., Zheng, H., Liu, C., Shi, J., Liu, X., and Lin, D.
\newblock Dsnas: Direct neural architecture search without parameter
  retraining.
\newblock \emph{2020 IEEE/CVF Conference on Computer Vision and Pattern
  Recognition (CVPR)}, pp.\  12081--12089, 2020.

\bibitem[Huang et~al.(2017)Huang, Liu, and Weinberger]{Huang2017DenselyCC}
Huang, G., Liu, Z., and Weinberger, K.~Q.
\newblock Densely connected convolutional networks.
\newblock \emph{2017 IEEE Conference on Computer Vision and Pattern Recognition
  (CVPR)}, pp.\  2261--2269, 2017.

\bibitem[Jiang \& Zhai(2007)Jiang and Zhai]{Jiang2007InstanceWF}
Jiang, J. and Zhai, C.
\newblock Instance weighting for domain adaptation in nlp.
\newblock In \emph{ACL}, 2007.

\bibitem[Kingma \& Ba(2015)Kingma and Ba]{Kingma2015AdamAM}
Kingma, D.~P. and Ba, J.
\newblock Adam: A method for stochastic optimization.
\newblock In \emph{ICLR}, 2015.

\bibitem[Krizhevsky(2009)]{Krizhevsky2009LearningML}
Krizhevsky, A.
\newblock Learning multiple layers of features from tiny images.
\newblock 2009.

\bibitem[Liang et~al.(2019)Liang, Zhang, Sun, He, Huang, Zhuang, and
  Li]{Liang2019DARTSID}
Liang, H., Zhang, S., Sun, J., He, X., Huang, W., Zhuang, K., and Li, Z.
\newblock Darts+: Improved differentiable architecture search with early
  stopping.
\newblock \emph{CoRR}, abs/1909.06035, 2019.

\bibitem[Lin et~al.(2017)Lin, Goyal, Girshick, He, and
  Doll{\'a}r]{Lin2017FocalLF}
Lin, T.-Y., Goyal, P., Girshick, R.~B., He, K., and Doll{\'a}r, P.
\newblock Focal loss for dense object detection.
\newblock \emph{2017 IEEE International Conference on Computer Vision (ICCV)},
  pp.\  2999--3007, 2017.

\bibitem[Liu et~al.(2018{\natexlab{a}})Liu, Zoph, Shlens, Hua, Li, Fei-Fei,
  Yuille, Huang, and Murphy]{Liu2018ProgressiveNA}
Liu, C., Zoph, B., Shlens, J., Hua, W., Li, L.-J., Fei-Fei, L., Yuille, A.~L.,
  Huang, J., and Murphy, K.~P.
\newblock Progressive neural architecture search.
\newblock In \emph{ECCV}, 2018{\natexlab{a}}.

\bibitem[Liu et~al.(2021)Liu, Haghgoo, Chen, Raghunathan, Koh, Sagawa, Liang,
  and Finn]{Liu2021JustTT}
Liu, E.~Z., Haghgoo, B., Chen, A.~S., Raghunathan, A., Koh, P.~W., Sagawa, S.,
  Liang, P., and Finn, C.
\newblock Just train twice: Improving group robustness without training group
  information.
\newblock In \emph{ICML}, 2021.

\bibitem[Liu et~al.(2018{\natexlab{b}})Liu, Simonyan, Vinyals, Fernando, and
  Kavukcuoglu]{Liu2018HierarchicalRF}
Liu, H., Simonyan, K., Vinyals, O., Fernando, C., and Kavukcuoglu, K.
\newblock Hierarchical representations for efficient architecture search.
\newblock 2018{\natexlab{b}}.

\bibitem[Liu et~al.(2019)Liu, Simonyan, and Yang]{Liu2019DARTSDA}
Liu, H., Simonyan, K., and Yang, Y.
\newblock Darts: Differentiable architecture search.
\newblock In \emph{ICLR}, volume abs/1806.09055, 2019.

\bibitem[Loshchilov \& Hutter(2017)Loshchilov and Hutter]{Loshchilov2017SGDRSG}
Loshchilov, I. and Hutter, F.
\newblock Sgdr: Stochastic gradient descent with warm restarts.
\newblock In \emph{ICLR}, 2017.

\bibitem[Ma et~al.(2018)Ma, Zhang, Zheng, and Sun]{Ma2018ShuffleNetVP}
Ma, N., Zhang, X., Zheng, H., and Sun, J.
\newblock Shufflenet v2: Practical guidelines for efficient cnn architecture
  design.
\newblock In \emph{ECCV}, 2018.

\bibitem[Mirza \& Osindero(2014)Mirza and Osindero]{Mirza2014ConditionalGA}
Mirza, M. and Osindero, S.
\newblock Conditional generative adversarial nets.
\newblock \emph{CoRR}, abs/1411.1784, 2014.

\bibitem[Moore \& Lewis(2010)Moore and Lewis]{Moore2010IntelligentSO}
Moore, R.~C. and Lewis, W.~D.
\newblock Intelligent selection of language model training data.
\newblock In \emph{ACL}, 2010.

\bibitem[Ngiam et~al.(2018)Ngiam, Peng, Vasudevan, Kornblith, Le, and
  Pang]{Ngiam2018DomainAT}
Ngiam, J., Peng, D., Vasudevan, V., Kornblith, S., Le, Q.~V., and Pang, R.
\newblock Domain adaptive transfer learning with specialist models.
\newblock \emph{CoRR}, abs/1811.07056, 2018.

\bibitem[Noy et~al.(2020)Noy, Nayman, Ridnik, Zamir, Doveh, Friedman, Giryes,
  and Zelnik-Manor]{Noy2020ASAPAS}
Noy, A., Nayman, N., Ridnik, T., Zamir, N., Doveh, S., Friedman, I., Giryes,
  R., and Zelnik-Manor, L.
\newblock Asap: Architecture search, anneal and prune.
\newblock In \emph{AISTATS}, 2020.

\bibitem[Pham et~al.(2018)Pham, Guan, Zoph, Le, and Dean]{Pham2018EfficientNA}
Pham, H., Guan, M.~Y., Zoph, B., Le, Q.~V., and Dean, J.
\newblock Efficient neural architecture search via parameter sharing.
\newblock In \emph{ICML}, 2018.

\bibitem[Radford et~al.(2016)Radford, Metz, and
  Chintala]{Radford2016UnsupervisedRL}
Radford, A., Metz, L., and Chintala, S.
\newblock Unsupervised representation learning with deep convolutional
  generative adversarial networks.
\newblock In \emph{ICLR}, 2016.

\bibitem[Real et~al.(2017)Real, Moore, Selle, Saxena, Suematsu, Tan, Le, and
  Kurakin]{Real2017LargeScaleEO}
Real, E., Moore, S., Selle, A., Saxena, S., Suematsu, Y.~L., Tan, J., Le,
  Q.~V., and Kurakin, A.
\newblock Large-scale evolution of image classifiers.
\newblock In \emph{ICML}, 2017.

\bibitem[Real et~al.(2019)Real, Aggarwal, Huang, and Le]{Real2019RegularizedEF}
Real, E., Aggarwal, A., Huang, Y., and Le, Q.~V.
\newblock Regularized evolution for image classifier architecture search.
\newblock In \emph{AAAI}, 2019.

\bibitem[Ren et~al.(2018)Ren, Zeng, Yang, and Urtasun]{Ren2018LearningTR}
Ren, M., Zeng, W., Yang, B., and Urtasun, R.
\newblock Learning to reweight examples for robust deep learning.
\newblock In \emph{ICML}, 2018.

\bibitem[Ren et~al.(2020)Ren, Yeh, and Schwing]{Ren2020NotAU}
Ren, Z., Yeh, R.~A., and Schwing, A.~G.
\newblock Not all unlabeled data are equal: Learning to weight data in
  semi-supervised learning.
\newblock In \emph{NeurIPS}, 2020.

\bibitem[Shu et~al.(2019)Shu, Xie, Yi, Zhao, Zhou, Xu, and
  Meng]{Shu2019MetaWeightNetLA}
Shu, J., Xie, Q., Yi, L., Zhao, Q., Zhou, S., Xu, Z., and Meng, D.
\newblock Meta-weight-net: Learning an explicit mapping for sample weighting.
\newblock In \emph{NeurIPS}, 2019.

\bibitem[Sivasankaran et~al.(2017)Sivasankaran, Vincent, and
  Illina]{Sivasankaran2017DiscriminativeIW}
Sivasankaran, S., Vincent, E., and Illina, I.
\newblock Discriminative importance weighting of augmented training data for
  acoustic model training.
\newblock \emph{2017 IEEE International Conference on Acoustics, Speech and
  Signal Processing (ICASSP)}, pp.\  4885--4889, 2017.

\bibitem[Such et~al.(2020)Such, Rawal, Lehman, Stanley, and
  Clune]{Such2020GenerativeTN}
Such, F.~P., Rawal, A., Lehman, J., Stanley, K.~O., and Clune, J.
\newblock Generative teaching networks: Accelerating neural architecture search
  by learning to generate synthetic training data.
\newblock In \emph{ICML}, 2020.

\bibitem[Szegedy et~al.(2015)Szegedy, Liu, Jia, Sermanet, Reed, Anguelov,
  Erhan, Vanhoucke, and Rabinovich]{Szegedy2015GoingDW}
Szegedy, C., Liu, W., Jia, Y., Sermanet, P., Reed, S.~E., Anguelov, D., Erhan,
  D., Vanhoucke, V., and Rabinovich, A.
\newblock Going deeper with convolutions.
\newblock \emph{2015 IEEE Conference on Computer Vision and Pattern Recognition
  (CVPR)}, pp.\  1--9, 2015.

\bibitem[Tan et~al.(2019)Tan, Chen, Pang, Vasudevan, and Le]{Tan2019MnasNetPN}
Tan, M., Chen, B., Pang, R., Vasudevan, V., and Le, Q.~V.
\newblock Mnasnet: Platform-aware neural architecture search for mobile.
\newblock \emph{2019 IEEE/CVF Conference on Computer Vision and Pattern
  Recognition (CVPR)}, pp.\  2815--2823, 2019.

\bibitem[Wang et~al.(2020{\natexlab{a}})Wang, Pham, Michel, Anastasopoulos,
  Neubig, and Carbonell]{Wang2020OptimizingDU}
Wang, X., Pham, H., Michel, P., Anastasopoulos, A., Neubig, G., and Carbonell,
  J.~G.
\newblock Optimizing data usage via differentiable rewards.
\newblock In \emph{ICML}, 2020{\natexlab{a}}.

\bibitem[Wang et~al.(2020{\natexlab{b}})Wang, Xue, Yan, Yang, Hu, and
  Sun]{Wang2020MergeNASMO}
Wang, X., Xue, C., Yan, J., Yang, X., Hu, Y., and Sun, K.
\newblock Mergenas: Merge operations into one for differentiable architecture
  search.
\newblock In \emph{IJCAI}, 2020{\natexlab{b}}.

\bibitem[Wang et~al.(2020{\natexlab{c}})Wang, Guo, Song, and
  Huang]{Wang2020MetaSemiAM}
Wang, Y., Guo, J., Song, S., and Huang, G.
\newblock Meta-semi: A meta-learning approach for semi-supervised learning.
\newblock \emph{CoRR}, abs/2007.02394, 2020{\natexlab{c}}.

\bibitem[Xie et~al.(2020)Xie, Du, and Ban]{Xie2020SkillearnML}
Xie, P., Du, X., and Ban, H.
\newblock Skillearn: Machine learning inspired by humans' learning skills.
\newblock \emph{ArXiv}, abs/2012.04863, 2020.

\bibitem[Xie et~al.(2019)Xie, Zheng, Liu, and Lin]{Xie2019SNASSN}
Xie, S., Zheng, H., Liu, C., and Lin, L.
\newblock Snas: Stochastic neural architecture search.
\newblock In \emph{ICLR}, 2019.

\bibitem[Xu et~al.(2020)Xu, Xie, Zhang, Chen, Qi, Tian, and
  Xiong]{Xu2020PCDARTSPC}
Xu, Y., Xie, L., Zhang, X., Chen, X., Qi, G.-J., Tian, Q., and Xiong, H.
\newblock Pc-darts: Partial channel connections for memory-efficient
  architecture search.
\newblock In \emph{ICLR}, 2020.

\bibitem[Zela et~al.(2020)Zela, Elsken, Saikia, Marrakchi, Brox, and
  Hutter]{Zela2020UnderstandingAR}
Zela, A., Elsken, T., Saikia, T., Marrakchi, Y., Brox, T., and Hutter, F.
\newblock Understanding and robustifying differentiable architecture search.
\newblock In \emph{ICLR}, 2020.

\bibitem[Zhang et~al.(2019)Zhang, Goodfellow, Metaxas, and
  Odena]{Zhang2019SelfAttentionGA}
Zhang, H., Goodfellow, I.~J., Metaxas, D.~N., and Odena, A.
\newblock Self-attention generative adversarial networks.
\newblock In \emph{ICML}, 2019.

\bibitem[Zhang et~al.(2018)Zhang, Zhou, Lin, and Sun]{Zhang2018ShuffleNetAE}
Zhang, X., Zhou, X., Lin, M., and Sun, J.
\newblock Shufflenet: An extremely efficient convolutional neural network for
  mobile devices.
\newblock \emph{2018 IEEE/CVF Conference on Computer Vision and Pattern
  Recognition}, pp.\  6848--6856, 2018.

\bibitem[Zhou et~al.(2019)Zhou, Yang, Wang, and Pan]{Zhou2019BayesNASAB}
Zhou, H., Yang, M., Wang, J., and Pan, W.
\newblock Bayesnas: A bayesian approach for neural architecture search.
\newblock In \emph{ICML}, pp.\  7603--7613, 2019.

\bibitem[Zoph \& Le(2017)Zoph and Le]{Zoph2017NeuralAS}
Zoph, B. and Le, Q.~V.
\newblock Neural architecture search with reinforcement learning.
\newblock In \emph{ICLR}, 2017.

\bibitem[Zoph et~al.(2018)Zoph, Vasudevan, Shlens, and Le]{Zoph2018LearningTA}
Zoph, B., Vasudevan, V., Shlens, J., and Le, Q.~V.
\newblock Learning transferable architectures for scalable image recognition.
\newblock pp.\  8697--8710, 2018.

\end{thebibliography}
\bibliographystyle{icml2022}

\newpage
\appendix
\onecolumn

\section{Visualization of generated images}
\label{app:visualization}

\begin{figure}[ht]
    \vskip 0.2in
    \begin{center}
    \includegraphics[width=0.45\columnwidth]{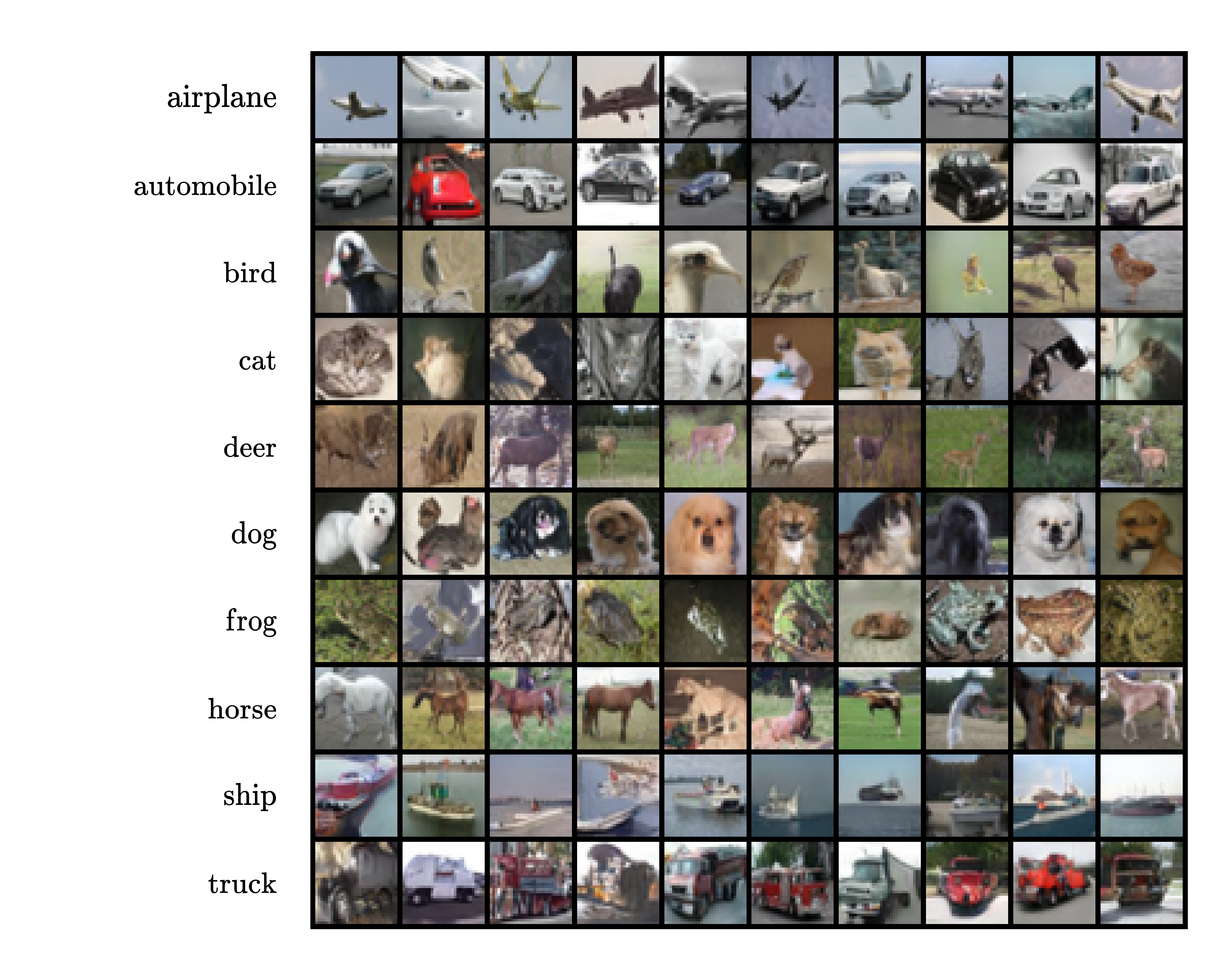}
    \hspace{2pt}
    \includegraphics[width=0.45\columnwidth]{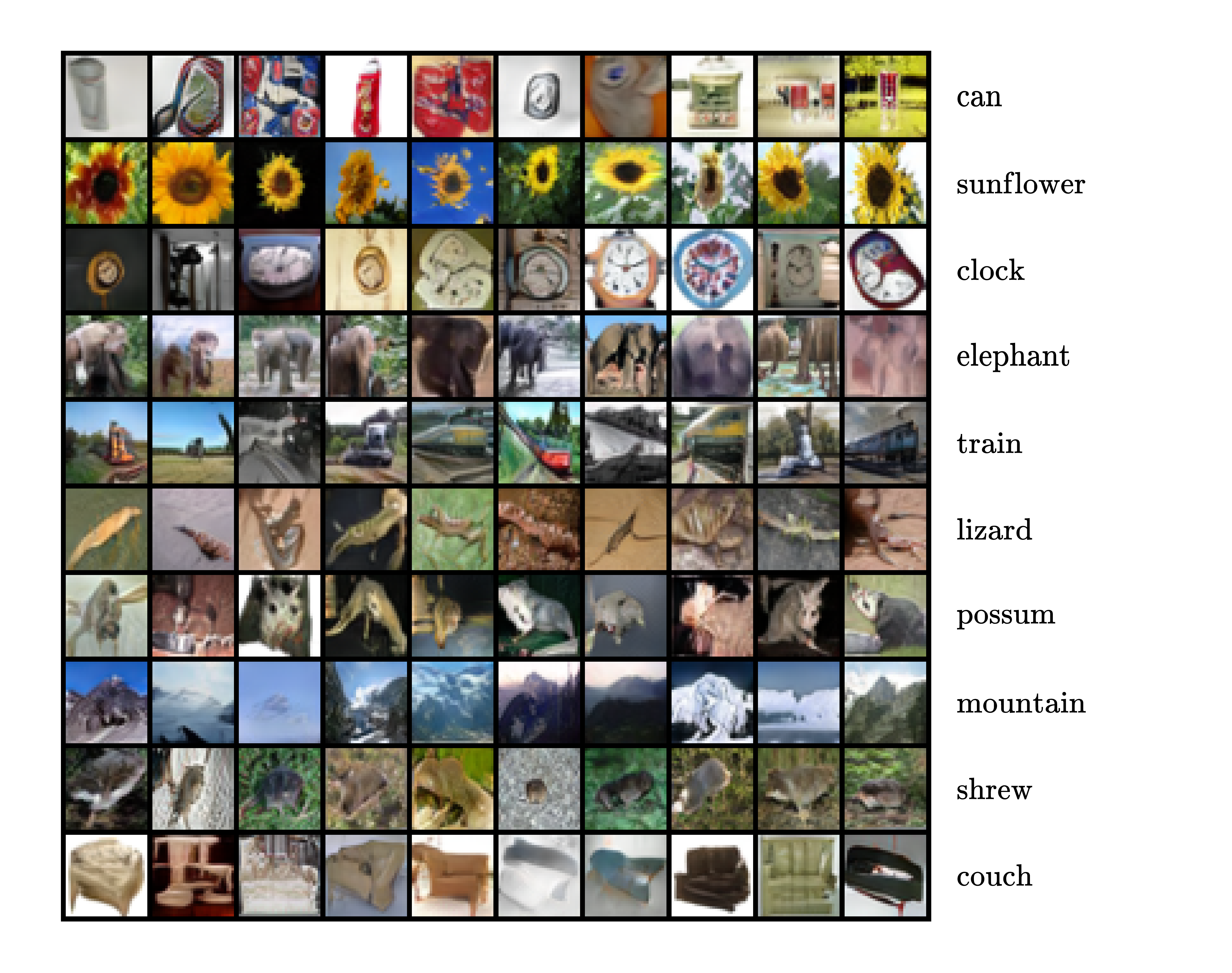}
    \caption{Samples generated from BigGAN conditioned on class distribution for CIFAR-10 (left) and CIFAR-100 (right) datasets during DARTS-1st search space.}
    \label{fig:cifar_samples}
    \end{center}
    \vskip -0.2in
\end{figure}
\begin{figure}[!ht]
    \vskip 0.2in
    \begin{center}
    \includegraphics[width=0.9\columnwidth]{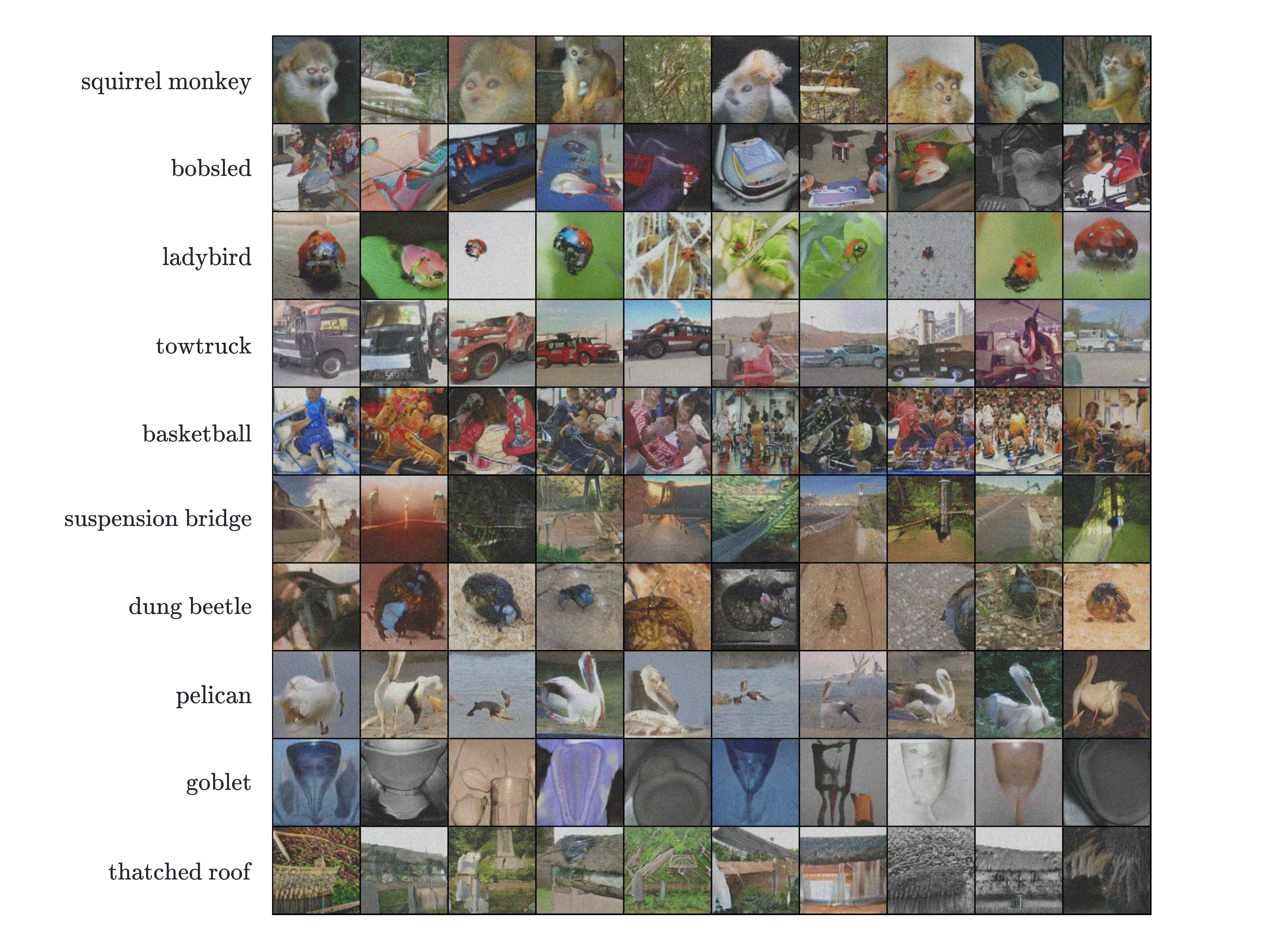}
    \caption{Images generated from BigGAN conditioned on class distribution for ImageNet dataset during PC-DARTS search space.}
    \label{fig:imagenet_samples}
    \end{center}
    \vskip -0.2in
\end{figure}

Figure \ref{fig:cifar_samples} and Figure \ref{fig:imagenet_samples} visualizes the sample of images generated by CIG (BigGAN) in the search phase for our approach with DARTS-1st and PC-DARTS, respectively.

\newpage
\section{Full lists of hyperparameter settings}
\label{app:params_settings}

Table \ref{tab:arch_search} and \ref{tab:arch_eval} show the hyperparameter settings used in different experiments in the search and evaluation phase.

\begin{table}[htb]
    \caption{Hyperparameter settings for our approach with DARTS, P-DARTS, and PC-DARTS on CIFAR and ImageNet datasets during architecture search. The rest of the parameters for $G$ and $H$ are used as described in BigGAN \cite{Brock2019LargeSG}.}
    \label{tab:arch_search}
    \vskip 0.15in
    \begin{center}
    \begin{small}
    \begin{tabular}{{p{4.6cm} p{1.8cm} p{1.8cm} p{1.8cm} p{1.8cm}}}
    \toprule
    Hyperparameters & DARTS\hspace{1em} (CIFAR) & P-DARTS\hspace{0.5em} (CIFAR) & PC-DARTS (CIFAR) & PC-DARTS (ImageNet) \\
    \midrule
    Optimizer for $W_1$,$W_2$ & SGD & SGD & SGD & SGD \\
    Initial learning rate for $W_1$,$W_2$ & 0.025 & 0.025 & 0.025 & 0.5 \\
    Learning rate scheduler for $W_1$,$W_2$ & Cosine decay & Cosine decay & Cosine decay & Cosine decay \\
    Minimum learning rate for $W_1$,$W_2$ & 0.001 & 0.001 & 0.001 & 0.005 \\
    Momentum for $W_1$,$W_2$ & 0.9 & 0.9 & 0.9 & 0.9 \\
    Weight decay for $W_1$,$W_2$ & 0.0003 & 0.0003 & 0.0003 & 0.0003 \\
    Optimizer for A & Adam & Adam & Adam & Adam \\
    Learning rate for A & 0.0003 & 0.0003 & 0.0003 & 0.006 \\
    Weight decay for A & 0.001 & 0.001 & 0.001 & 0.001 \\
    Optimizer for $G$,$H$ & Adam & Adam & Adam & Adam \\
    Learning rate for $G$,$H$ & 0.0002 & 0.0002 & 0.0002 & 0.0002 \\
    Initial channels for $W_1$,$W_2$ & 16 & 16 & 16 & 16 \\
    Layers for $W_1$,$W_2$ & 8 & 8 & 8 & 8 \\
    Gradient Clip for $W_1$,$W_2$ & 5 & 5 & 5 & 5 \\
    Batch size & 64 & 64 & 64 & 64 \\
    Epochs & 50 & 25 & 50 & 50 \\
    Add layers & - & [6,12] & - & - \\
    Dropout rate & - & [0.1,0.4,0.7] & - & - \\
    $\lambda$ & 1 & 1 & 1 & 1 \\
    \bottomrule
    \end{tabular}
    \end{small}
    \end{center}
    \vskip -0.1in
\end{table}

\begin{table}[htb]
    \caption{Hyperparameter settings for our approach with DARTS, P-DARTS, and PC-DARTS on CIFAR and ImageNet datasets during architecture evaluation.}
    \label{tab:arch_eval}
    \vskip 0.15in
    \begin{center}
    \begin{small}
    \begin{tabular}{{p{3cm} p{1.7cm} p{1.7cm} p{1.7cm} p{1.7cm} p{1.7cm} p{1.7cm}}}
    \toprule
    Hyperparameters & DARTS\hspace{1em} (CIFAR) & P-DARTS (CIFAR) & PC-DARTS (CIFAR) & DARTS\hspace{1em} (ImageNet) & P-DARTS (ImageNet) & PC-DARTS (ImageNet) \\
    \midrule
    Optimizer & SGD & SGD & SGD & SGD & SGD & SGD \\
    Initial learning rate & 0.025 & 0.025 & 0.025 & 0.5 & 0.5 & 0.5 \\
    Learning rate scheduler & Cosine decay & Cosine decay & Cosine decay & Cosine decay & Cosine decay & Cosine decay \\
    Momentum & 0.9 & 0.9 & 0.9 & 0.9 & 0.9 & 0.9 \\
    Weight decay & 0.0003 & 0.0003 & 0.0003 & 0.00003 & 0.00003 & 0.00003 \\
    Initial channels & 36 & 36 & 36 & 48 & 48 & 48 \\
    Layers & 20 & 20 & 20 & 14 & 14 & 14 \\
    Auxiliary weight & 0.4 & 0.4 & 0.4 & 0.4 & 0.4 & 0.4 \\
    Cutout length & 16 & 16 & 16 & - & - & - \\
    Label smooth & - & - & - & 0.1 & 0.1 & 0.1 \\
    Drop path prob & 0.3 & 0.3 & 0.3 & 0.0 & 0.0 & 0.0 \\
    Gradient Clip & 5 & 5 & 5 & 5 & 5 & 5 \\
    Batch size & 96 & 96 & 96 & 1024 & 1024 & 1024 \\
    Epochs & 600 & 600 & 600 & 250 & 250 & 250 \\
    \bottomrule
    \end{tabular}
    \end{small}
    \end{center}
    \vskip -0.1in
\end{table}


\end{document}